\documentclass[10pt,twocolumn,letterpaper]{article}

\usepackage[pagenumbers]{cvpr} 

\usepackage{graphicx}
\usepackage{amsmath}
\usepackage{amssymb}
\usepackage{booktabs}
\usepackage{ulem}

\usepackage[pagebackref,breaklinks,colorlinks]{hyperref}
\usepackage[capitalize]{cleveref}
\crefname{section}{Sec.}{Secs.}
\Crefname{section}{Section}{Sections}
\Crefname{table}{Table}{Tables}
\crefname{table}{Tab.}{Tabs.}

\def\cvprPaperID{10281} 
\def\confName{CVPR}
\def\confYear{2023}

\usepackage{xcolor}         
\usepackage{pifont}

\definecolor{myred}{RGB}{220,50,47} 
\definecolor{mygreen}{RGB}{133,153,0}
\definecolor{commentcolor}{RGB}{133,153,0}
\definecolor{urlcolor}{rgb}{0.93,0.01,0.55}

\newcommand*{\affmark}[1][*]{\textsuperscript{#1}}

\usepackage{mathtools}

\newcommand{\E}{\mathbb{E}}

\newcommand{\Eb}[2]{\E_{#1}\!\left[#2\right]}

\newcommand{\bI}{\mathbf{I}}

\newcommand{\bzero}{\mathbf{0}}

\newcommand{\bc}{\mathbf{c}}

\newcommand{\bx}{\mathbf{x}}

\newcommand{\bz}{\mathbf{z}}

\newcommand{\bepsilon}{{\boldsymbol{\epsilon}}}

\usepackage{enumitem}

\begin{document}

\title{Diffusion Guided Domain Adaptation of Image Generators}


\author{Kunpeng Song\affmark[1]\quad\quad Ligong Han\affmark[1]\quad\quad Bingchen Liu\affmark[2] \quad\quad
Dimitris Metaxas\affmark[1] \quad\quad Ahmed Elgammal\affmark[1, 3]\\
{\affmark[1]Rutgers University\quad\quad\quad\affmark[2]Bytedance Inc.\quad\quad\quad\affmark[3]Playform AI}
}

\maketitle

\begin{abstract}
   Can a text-to-image diffusion model be used as a training objective for adapting a GAN generator to another domain? In this paper, we show that the classifier-free guidance can be leveraged as a critic and enable generators to distill knowledge from large-scale text-to-image diffusion models. 
   Generators can be efficiently shifted into new domains indicated by text prompts without access to groundtruth samples from target domains. We demonstrate the effectiveness and controllability of our method through extensive experiments. Although not trained to minimize CLIP loss, our model achieves equally high CLIP scores and significantly lower FID than prior work on short prompts, and outperforms the baseline qualitatively and quantitatively on long and complicated prompts. To our best knowledge, the proposed method is the first attempt at incorporating large-scale pre-trained diffusion models and distillation sampling for text-driven image generator domain adaptation and gives a quality previously beyond possible. Moreover, we extend our work to 3D-aware style-based generators and DreamBooth guidance. For code and more visual samples, please visit our~\href{https://styleganfusion.github.io/}{\color{urlcolor}{Project Webpage}}.
\end{abstract}

\begin{figure*}
\centering
\includegraphics[width=1.0\linewidth]{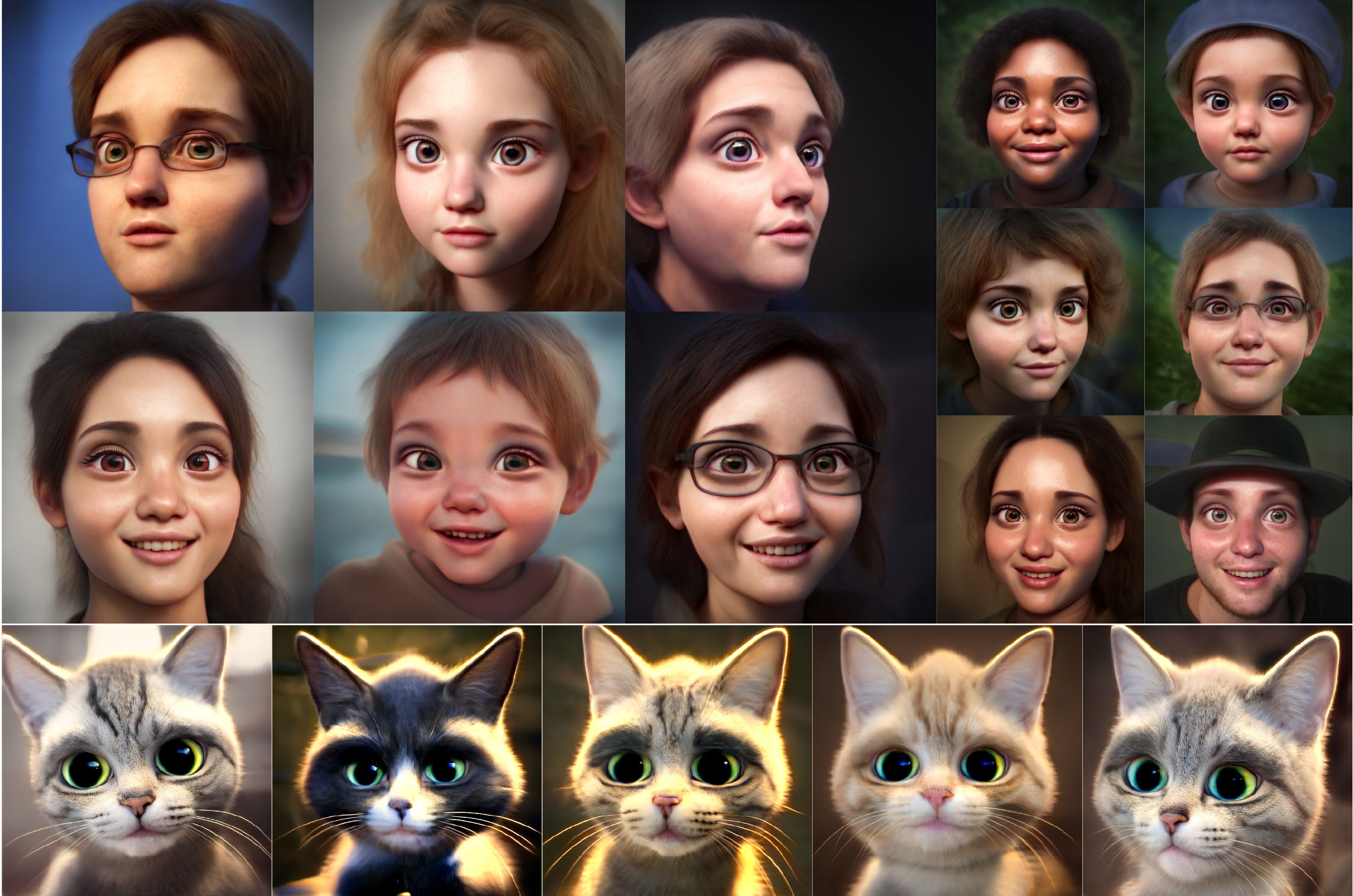}
  \caption{Example images after adapting generator to a domain specified by a text description. The first section is a photo from the FFHQ dataset to 3D stylized Anime, the second section is from cats to 3D rendering cats. Detailed text prompts can be founded in the appendix.
  }
  \label{fig:hero}
\end{figure*}

\section{Introduction}\label{sec:intro}
Diffusion models have witnessed a remarkable rise in image-generation tasks with their ability to cover a wide range of visual semantics from different image domains~\cite{ramesh2021zero,ding2021cogview,nichol2021glide, rombach2021high,ramesh2022hierarchical,saharia2022photorealistic,yu2022scaling}. These models usually require lots of iterative generative steps, which is computationally demanding and makes it undesirable in many application scenarios. On the other hand, GANs~\cite{goodfellow2014generative,karras2017progressive,karras2019style,karras2020analyzing,karras2021alias} preserve their advantage over diffusion models in terms of generation speed and is more computational-friendly to train. Within a single image domain, GANs can have an accessible latent space with the expressive power to synthesize images with fine-grained variations. Leveraging a pre-trained text-to-image diffusion model, such as StableDiffusion, we propose a new training objective that can quickly shift a pre-trained GAN model into another image domain. We take advantage of StableDiffusion's prior knowledge learned from enormous text-and-image pairs and use such prior guiding the GAN model to shift its generation behavior. With the developed training objective, are able to shift the output of GAN to a totally different image domain without the need for any training images in that domain.

The availability of large-scale text-to-image models unleashes the potential of zero-shot domain shifting for GANs. Prior works, such as StyleGAN-NADA~\cite{gal2022stylegan}, take advantage of CLIP’s~\cite{radford2021learning} power to relate visual features to textual semantics. Via a quick training of minimizing CLIP’s image-to-text similarity on a certain set of prompts, StyleGAN-NADA is able to generate cartoon avatars  from a model trained only on realistic faces and oil paintings from a model initially trained only on photographs. However, these methods rely on CLIP, which has been shown to proceed with a misaligned text and image latent space~\cite{Liang2022MindTG}. CLIP loss is known hard to be minimized in previous research\cite{liu2021fusedream} as it tends to be trapped in a local minimum. This limits the effectiveness of the selected prompts to drive the GAN toward the desired image domain and can lead to image artifacts and hurts the generation diversity of the original GAN model. 

In this work, we explore using diffusion to improve the performance of a text-driven image generator for domain adaptation. We leverage the power of pre-trained large-scale diffusion models and build on the recently proposed Score Distillation Sampling technique~\cite{poole2022dreamfusion}, where text-to-image diffusion acts as a frozen, efficient critic that predicts image-space edits. Our new domain adaptation method takes advantage of a pre-trained image diffusion model, providing well-aligned guidance directly from the image domain to help train the GAN model.

Intuitively, distilling from a generative model (such as a diffusion model) can provide more informative signals for a generator than a discriminative model (such as CLIP). In this paper, we investigate two techniques that combine diffusion with style-based generators to explore this idea further:
\begin{itemize}[noitemsep,nolistsep]
    \item We introduce the diffusion model score distillation sampling (SDS) into domain adaptation of style-based image generators and achieve better performance than the prior art. 
    \item To regularize the network and prevent model collapse, we propose a diffusion directional regularizer and adapt the reconstruction guidance to SDS. To solve blurry issues, we adapt the layer selection into the SDS finetuning framework.
\end{itemize}


\section{Related Work} 

\noindent \textbf{Image generator domain adaptation.}
How can we get a generator without having access to enough real data? The goal of domain adaptation is to shift the data distribution of image generators to a desired new domain different from what it is trained on. Prior works branches into two directions: few-shot and text-guided zero-shot fine-tuning. 

Few-shot models are trained with several hundred or fewer\cite{ojha2021few} image samples. To better capture the target domain, some control channel statistics \cite{noguchi2019image} or sampling process \cite{wang2020minegan} in the latent space. Regularizer to prevent model collapse issue~\cite{mo2020freeze,pinkney2020resolution,robb2020few,li2020few,tseng2021regularizing}. 
Some use auxiliary tasks~\cite{liu2020towards,yang2021data} to alleviate overfitting. 
Text-guided zero-shot fine-tuning uses only text as guidance. Prior works exploited the semantic power of large-scale CLIP models~\cite{radford2021learning} to find editable latent space directions in a pre-trained StyleGAN2~\cite{karras2019analyzing}. For example, StyleCLIP~\cite{patashnik2021styleclip} optimizes the latent code for the generator and minimizes the text-image similarity score from CLIP. StyleGAN-NADA~\cite{gal2022stylegan} takes a step further by directly fine-tuning the generator using the CLIP text-image directional objective. 

More recently, diffusion models show great potential in text-guided fine-tuning. similar to StyleCLIP\cite{patashnik2021styleclip},  DiffusionCLIP~\cite{kim2022diffusionclip} applies CLIP~\cite{radford2021learning} objective to diffusion generators. Some fine-tune the text embedding~\cite{gal2022image} or the full diffusion model~\cite{ruiz2022dreambooth} on a few personalized images. 

\noindent \textbf{Text to image diffusion model.} 
Diffusion models~\cite{sohl2015deep,ho2020denoising} have achieved state-of-the-art image synthesis quality~\cite{song2021denoising,nichol2021glide,saharia2022photorealistic,preechakul2022diffusion}, especially on large-scale text-to-image synthesis tasks. Introduced by~\cite{ho2020denoising}, diffusion models use an iterative denoising process, which enables them to iteratively convert Gaussian noise into fine-grained images from a diverse and complicated image distribution.
Latent diffusion models (LDMs)~\cite{rombach2022high} is a class of diffusion models that operates on a latent space of a pre-trained autoencoder. Instead of learning directly from the image space, learning from the latent space greatly reduces the data sample dimension. The latent space comes with well-compressed semantic features and visual patterns that are already learned by the autoencoder, thus saving the cost of the diffusion model to learn everything from scratch. StableDiffusion~\cite{rombach2022high} is a popular representative of the latent diffusion models

\noindent \textbf{Score distillation sampling.} Prior works use diffusion models as critics to optimize an image or a Differentiable Image Parameterization (DIP)~\cite{mordvintsev2018differentiable} and bring it toward the distribution indicated by a text prompt. DreamFusion~\cite{poole2022dreamfusion} is a recent work that proposed a Score Distillation Sampling (SDS) loss to utilize a pre-trained text-to-image diffusion model~\cite{saharia2022photorealistic} to guide the training of NeRF~\cite{mildenhall2021nerf}. Their proposed  method can efficiently bypass the score-predicting module and approximate the gradient with the difference between the classifier-free guidance score and the ground-truth noise. DreamFusion~\cite{poole2022dreamfusion} performs SDS on image pixel space. We adopt the same gradient trick but extend it to the latent space of StableDiffusion~\cite{rombach2022high} and use it as guidance for StyleGAN2~\cite{karras2019analyzing} generator domain adaptation. We also include experiments with 3D-aware domain adaptation on EG3D~\cite{chan2022efficient} image generators. 

\section{Methods}
\begin{figure*}
\centering
\includegraphics[width=0.9\linewidth]{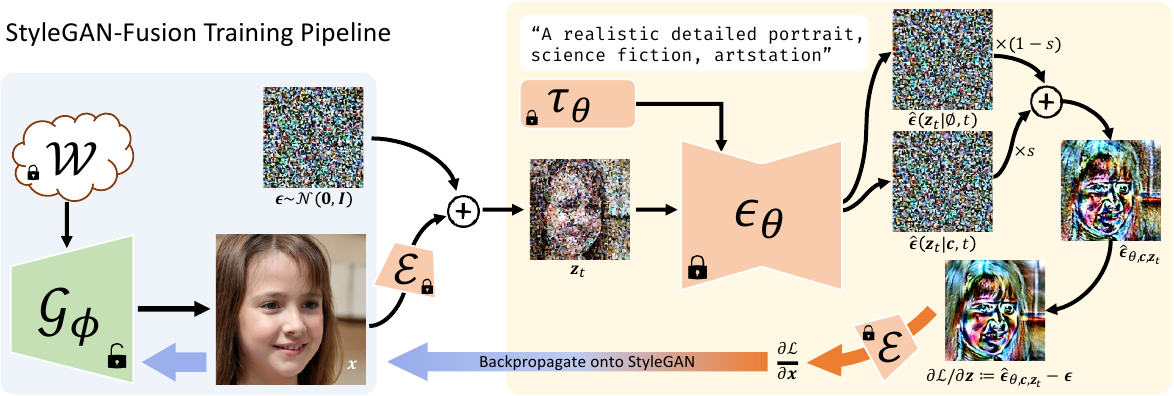}
  \caption{Overview of our StyleGAN-Fusion framework. The style-based generator $\mathcal{G}_\phi$ receives the gradient $\frac{\partial\mathcal{L}}{\partial \bx}$ backpropagated from $\frac{\partial\mathcal{L}}{\partial \bz}$ through encoder $\mathcal{E}$. $\hat{\bepsilon}_{\theta,\bc,\bz_t}$ is the classifier-free guidance score. All noises and noisy images are the decoded corresponding latents for visualization purposes.}
  \label{fig:pipeline}
\end{figure*}
\subsection{Background}
\noindent \textbf{Latent diffusion model.}
We use the publicly available latent diffusion model(LDM) StableDiffusion~\cite{rombach2022high} in this paper as our guidance model.
A LDM encodes images $\bx$ into latent space $\bz$ with an encoder $\mathcal{E}$, $\bz_0=\mathcal{E}(\bx)$, and the denoising process is preformed in the latent space $\mathcal{Z}$. Briefly, a latent diffusion model $\epsilon_\theta$ can be trained on a denoising objective of the following form:
\begin{equation}
\Eb{\bz_0,\bc,\bepsilon,t}{w_t \|\epsilon_\theta(\bz_t | \bc, t) - \bepsilon \|^2_2}
\label{eq:ldm}
\end{equation}
where $(\bx, \bc)$ are data-conditioning pairs, $\bepsilon \sim \mathcal{N}(\bzero, \bI)$, $t \sim \text{Uniform}(1, T)$ and $w_t$ is a weighting term.

\noindent \textbf{Classifier-free guidance.}
In the denoising sampling process, \textit{Classifier guidance} is an effective method to guide the synthesis better toward the desired direction, \textit{e.g.} a class or a text prompt~\cite{dhariwal2021diffusion}. The method uses gradients from a pre-trained model $p(\bc|\bz_t)$ during sampling.
\textit{Classifier-free guidance}  (CFG)~\cite{ho2022classifier} is an alternative technique that avoids this pre-trained classifier. During the training of the conditional diffusion model, randomly dropping the condition~$\bc$ lets the model learns to generate an image even without a condition. Therefore, a well-conditioned image can be generated by pushing the synthetic results under condition~$\bc$ further away from the unconditioned results during the diffusion process, where
\begin{align}
    \hat{\bepsilon}_{\theta,\bc}(\bz_t) = s \cdot \epsilon_\theta(\bz_t | \bc, t) + (1-s) \cdot \epsilon_{\theta}(\bz_t | \emptyset, t). \label{eq:classifier_free_score}
\end{align}
Here, $\epsilon_\theta(\bz_t | \bc, t)$ and $\epsilon_{\theta}(\bz_t | \emptyset, t)$ are conditional and unconditional $\bepsilon$-predictions. $s$ is the guidance weight and increasing $s > 1$ strengthens the effect of guidance. 

\subsection{Model Structure and Diffusion Guidance Loss}
An image $\mathbf{x}$ is generated with generator $\mathcal{G}$ from a style code $\mathbf{w} \sim P_\mathbf{w}$, where $P_\mathbf{w}$ is the pushforward measure from $\mathbf{z} \sim N(\mathbf{0},\mathbf{I})$ to $\mathcal{W}$ through a mapping network $g$. The generated image $\mathbf{x}=\mathcal{G}(\mathbf{w})$ is then encoded into the latent space of the StableDiffusion model using its encoder $\mathcal{E}$, $\mathbf{z}_0=\mathcal{E}(\mathbf{x}) \in \mathbb{R}^{c \times h \times w}$. Following the standard diffusion training schema, we sample a time-step $t \sim \text{Uniform}(1, T_x)$, with $0<T_x<T$, and perform the forward process (namely, ``q sample'') to get a noisy latent: $\mathbf{z}_t=q(\mathbf{z}_0,t):=\sqrt{\bar{\alpha}_t} \mathbf{z}_0 + \sqrt{1-\bar{\alpha}_t}\bepsilon$, with $\bepsilon \sim N(\mathbf{0},\mathbf{I})$.

Then the denoising process takes in $\mathbf{z}_t$ and returns the predicted noise $\hat{\bepsilon}_{\theta,\bc}(\bz_t)$ (classifier-free guidance score), conditioned on time step $t$ and text prompt embedding $y$. Ideally, if $\mathbf{z}_t$ is faithfully rendered according to the given text condition, the diffusion model $\epsilon_\theta$ should be able to correctly recover the true noise $\bepsilon$. We follow the gradient trick proposed by DreamFusion~\cite{poole2022dreamfusion} to directly used the difference between the predicted and the ground-truth scores as gradient and backpropagated through $\mathcal{E}$ to
the generator $\mathcal{G}$, $\nabla_\phi{\mathcal{G}_\phi}=\nabla_\bz{\mathcal{L}_{SDS}}\frac{\partial \bz}{\partial \phi}$, and
\begin{align}
  \nabla_\bz{\mathcal{L}_{SDS}}:=\Eb{\bc,\bepsilon,t}{w_t \left(\hat{\bepsilon}_{\theta,\bc}(\bz_t) - \bepsilon \right)}.
  \label{eq:important}
\end{align}
\noindent The generator parameters are updated accordingly. 
\subsection{Directional and Reconstruction Regularizer}
The diffusion guidance loss provides the generator with an informative direction to evolve, improving the image fidelity at the cost of diversity. It does not encourage image diversities. In fact, we observe that after a sufficient amount of iterations, the generator will collapse to a fixed image pattern regardless of its input noise. We assume this happens because the loss is sufficiently minimized when the unconditional $\hat{\bepsilon}_\emptyset$ is equal to the conditional $\hat{\bepsilon}_\mathbf{c}$ for all $\mathbf{w}$. It is searching for an image $\mathbf{x}$ that makes diffusion $\epsilon_\theta$ to predict the same noise for both unconditional and conditional inputs. This phenomenon is also known as differentiable image parameterization (or DIP). In such a case, the generator will lose image diversities, leading to a mode collapse. 

To address this issue, we regularize the generator optimization process with an additional loss term, which we defined as a diffusion directional regularizer. Denote the original frozen generator as $\mathcal{G}_{frozen}$ and the current one $\mathcal{G}_{train}$,
the classifier-free guidance scores are given by,
\begin{align}
    \hat{\bepsilon}_{train} &= \hat{\bepsilon}_{train,\emptyset} + s (\hat{\bepsilon}_{train,\mathbf{c}}-\hat{\bepsilon}_{train,\emptyset}) \nonumber\\
    \hat{\bepsilon}_{frozen} &= \hat{\bepsilon}_{frozen,\emptyset} + s (\hat{\bepsilon}_{frozen,\mathbf{c}}-\hat{\bepsilon}_{frozen,\emptyset})
\end{align}
The proposed directional regularizer is the cosine similarity between $\hat{\bepsilon}_{train}$ and $\hat{\bepsilon}_{frozen}$, maintaining a low directional difference. To efficiently implement it, we leverage the fact that a high dimensional Gaussian random variable lies on a sphere with high probability~\cite{huszar_2017} and minimize their $L_2$ distance instead. To do so, we normalize each score tensor according to its expected radius $r=\sqrt{c\times h \times w}$, and add a regularization gradient term defined as,
\begin{align}
    \nabla_{\bz}\mathcal{L}_{SDS}^{dir}:=r\left(\frac{\hat{\bepsilon}_{train}}{||\hat{\bepsilon}_{train}||_2}-\frac{\hat{\bepsilon}_{frozen}}{||\hat{\bepsilon}_{frozen}||_2}\right).
\end{align}

We use $\mathcal{L}_{SDS}^{dir}$ as a constraint on the optimization of the generator. Note that during the fine-tuning, $\hat{\bepsilon}_{frozen} $ is a fixed starting point for a given style code. If the generator only gives a single image for all $\mathbf{w}$, the gradient from all fixed starting points will be different. Such regularizer encourages $\mathcal{G}_{train}(\mathbf{w})$ to maintain its initial optimization direction, adding a force in preventing model collapse. Experiments show the directional regularizer can efficiently prevent model collapse. It is a plug-in module that is compatible with other regularization methods.

Additionally, we extend the score distillation framework to \textit{reconstruction guidance}~\cite{ho2022video} and introduce a reconstruction regularization. Intuitively, we want the current estimation of the clean latent image $\hat{\bz}_0=(\bz_t-\sqrt{1-\bar{\alpha}_t}\hat{\bepsilon})/\sqrt{\bar{\alpha}_t}$, given by the Tweedie's formula~\cite{chung2022diffusion}, to be similar to the latent image $\bz_0$ given by $\mathcal{G}_{frozen}$,
\begin{align}
    \nabla_{\bz}\mathcal{L}_{SDS}^{rec}:=r\left(\frac{\hat{\bepsilon}_{train}}{||\hat{\bepsilon}_{train}||_2}-\frac{\nabla_{\hat{\bepsilon}}\mathcal{L}_{rec}}{||\nabla_{\hat{\bepsilon}}\mathcal{L}_{rec}||_2}\right),
\end{align}
where $\nabla_{\hat{\bepsilon}}\mathcal{L}_{rec}$ is the gradient of reconstruction loss $\mathcal{L}_{rec}=\|\hat{\bz}_0-\bz_0\|_2^2$. We provide comparisons between the regularizers in the experiment section. The overall loss function is $\mathcal{L}=\mathcal{L}_{SDS}+\lambda_{dir}\mathcal{L}_{SDS}^{dir}+\lambda_{rec}\mathcal{L}_{SDS}^{rec}$, with $\lambda$'s the weighting coefficients.

\subsection{Timestep Range and Layer Selection}
The range of denoising timestep ($T_{SDS}$) from which timestep $t \sim \text{Uniform}(T_{min}, T_{max})$ is sampled closely relates to the model behavior. A larger $T_{SDS}$ results in a noisier latent after $q$ sample, leaving more room to the $\epsilon_\theta$ for modification. The guidance from $\epsilon_\theta$ is thus more related to high-level and overall image structures. A smaller $T_{SDS}$, on the other hand, leaves less room for guidance and is more related to local structures and details. Denoising timestep range configuration allows us to control the scale of changes (see \cref{T and layer exp}). 

Recall that style-based generators has a similar property: deeper layers control image composition and shallower layers the image details. Intuitively, if we optimize generator layers altogether, unsatisfied scenarios could occur where a high-level overall-structure guidance loss is used to update a shallow and detailed generator layer, resulting in blurry generated images. We use layer selection to overcome such issues. Inspired by StyleGAN-NADA~\cite{gal2022stylegan}, we perform $N$ iterations of optimization on the $\mathcal{W}^+$ style code space based on the SDS objective and select $k$ layers that correspond to the most significantly changed style codes. Ablation study (see \cref{T and layer exp}) shows the quality boost of multiple $k$ settings, especially in terms of reducing blurry vagueness. 


\section{Experiments}
\subsection{Results}

\begin{figure*}
\centering
\includegraphics[width=1.0\linewidth]{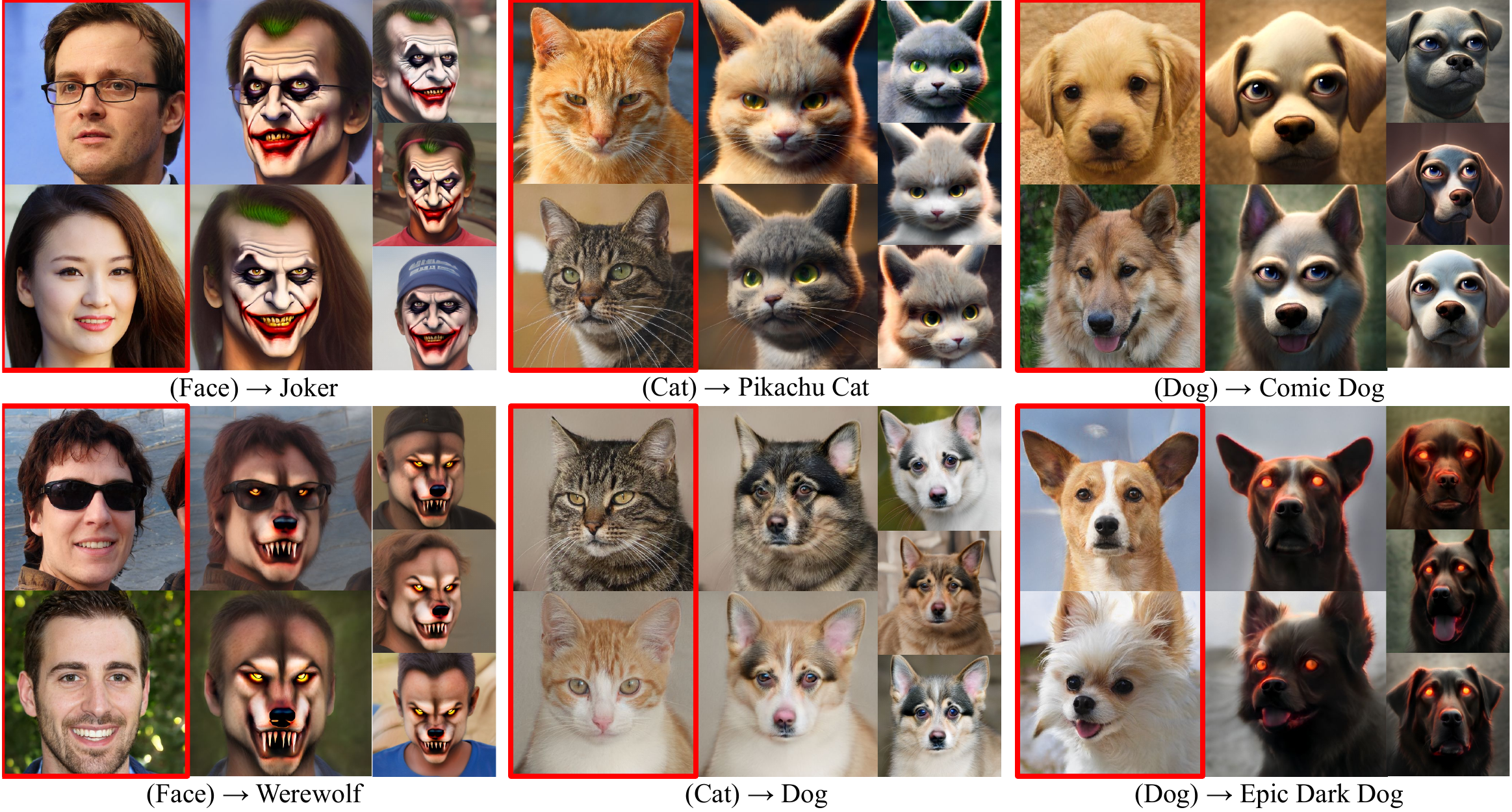}
  \caption{Generated images from experiments on FFHQ face, AFHQ-Cat and Dog~\cite{DBLP:journals/corr/abs-1912-01865}. The text below each section is the driving prompt. Notice our model only takes in a target prompt and does not need the source prompt. 
  }
  \label{fig:results_all}
\end{figure*}

We begin by showing a few result images. In \cref{fig:hero}, the upper section contains generated face images in a 3D rendering style described by the text promt. we take a StyleGAN2 generator pre-trained on the photorealistic FFHQ dataset and fine-tune it using our method. The lower section contains generated cat images, fine-tuned from the AFHQ-Cat~\cite{DBLP:journals/corr/abs-1912-01865} checkpoint. \cref{fig:results_all} shows more results including photo-realistic cars, werewolves, Joker, and photorealistic or 3D rendering of cat and dogs.
Full text prompts in each experiment and additional large-scale image galleries are included in the supplementary. 
We further extend our work to 3D experiments on EG3D~\cite{chan2022efficient} generators (see~\cref{sec:3D_ext}) and DreamBooth~\cite{ruiz2022dreambooth} guidance (see~\cref{sec:dreambooth_ext}).

\begin{figure}
\centering
\includegraphics[width=1.0\linewidth]{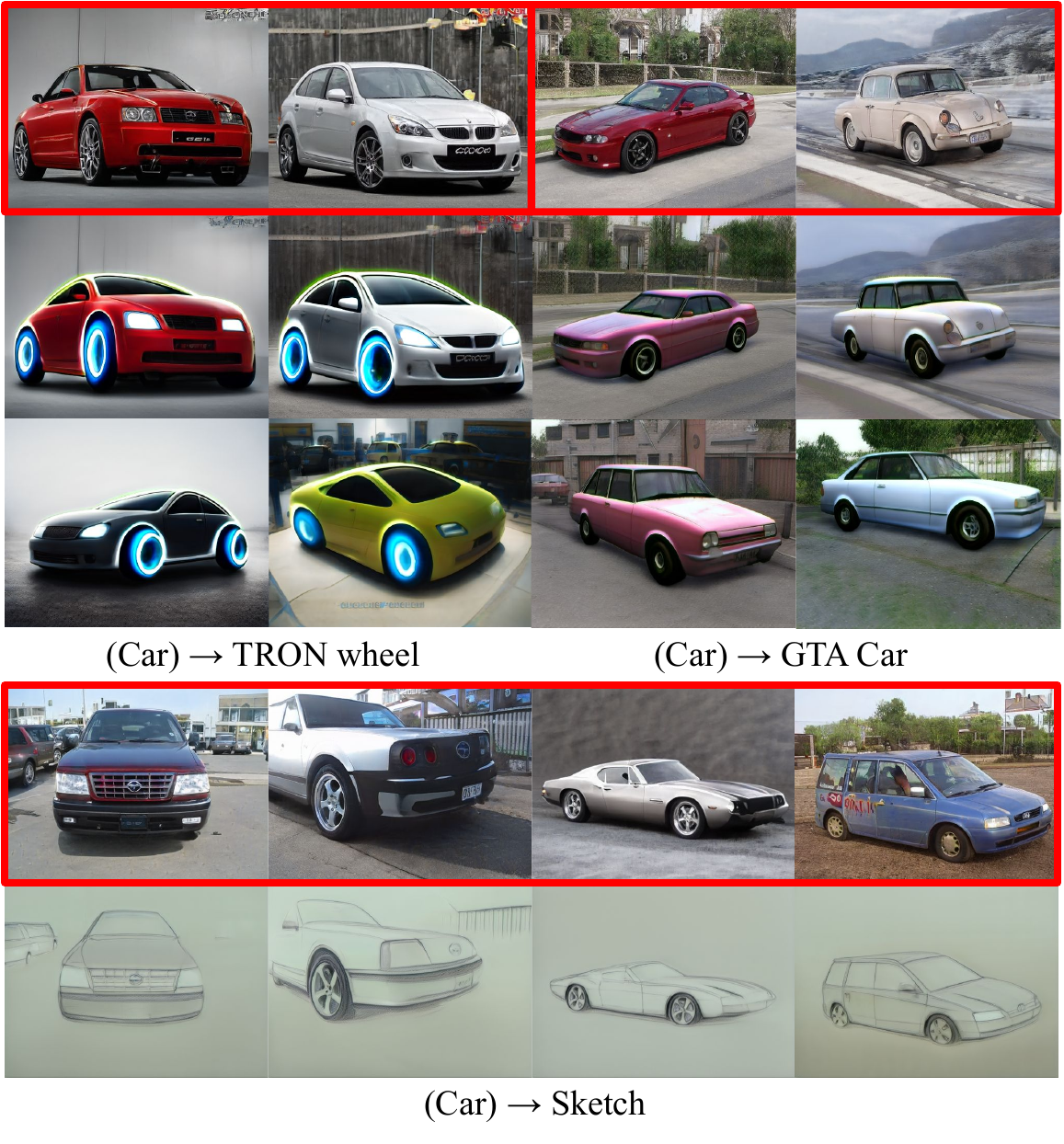}
  \caption{Generated image after adapting StyleGAN2-Car~\cite{NEURIPS2019_9015} model to new domains indicated by the prompts. 
  }
  \label{fig:GTA}
\end{figure}

\subsection{Comparison with Baseline}
\subsubsection{Long Text Prompts}
We compare our method and the baseline, StyleGAN-NADA, in multiple experiments with long text prompts which contain a detailed description of the desired image domain. Experiments show that the baseline has difficulty capturing all key constraints mentioned in the long text prompt. Our model, in contrast, generates images with significantly higher quality and fidelity when the text prompts are long and complicated. In this section, we show the quality difference from both visual and quantitative perspectives.

\begin{figure}
\centering
\includegraphics[width=1.0\linewidth]{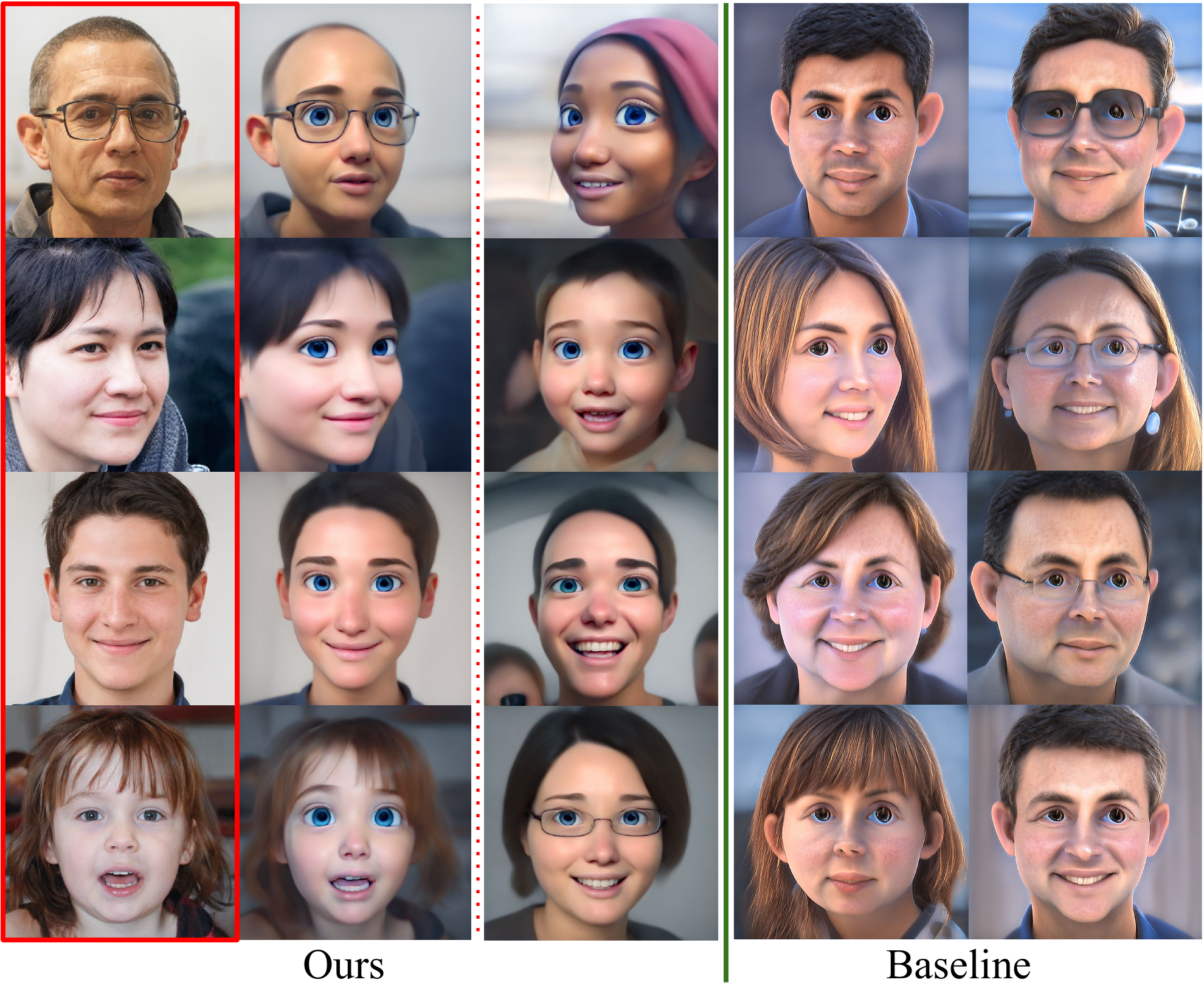}
  \caption{Compare our method with the baseline on "FFHQ face to 3D-stylized face". The text prompt is long, requesting a Pixar rendering with a cinematic smooth texture, reflective eyes, and 3D lighting. Notice the eyes in the baseline are not as big, beautiful, and reflective as ours. The baseline also contains unexpected textures, conflicting with the requirement. Our results better match the prompt and have more realistic 3D lighting.}
  \label{fig:nada_2}
\end{figure}

We conduct experiments on the FFHQ face model with long prompts that requires realistic 3D renderings and complicated lighting styles. \cref{fig:nada_2} shows the results from the baseline and ours. Baseline images have distorted shapes, especially eyes, and an oil-painting-like style, which conflicts with the smooth keyword. This happens because the limited information capacity of the CLIP sentence embedding fails to properly take care of a long text prompt. Our results better match the prompt, especially in terms of natural and undistorted face layouts and large beautiful reflective eyes.  Moreover, results from our model have more realistic 3D lighting and better match the text prompts. The full-text prompt is available in the appendix.

\begin{figure}
\centering
\includegraphics[width=1.0\linewidth]{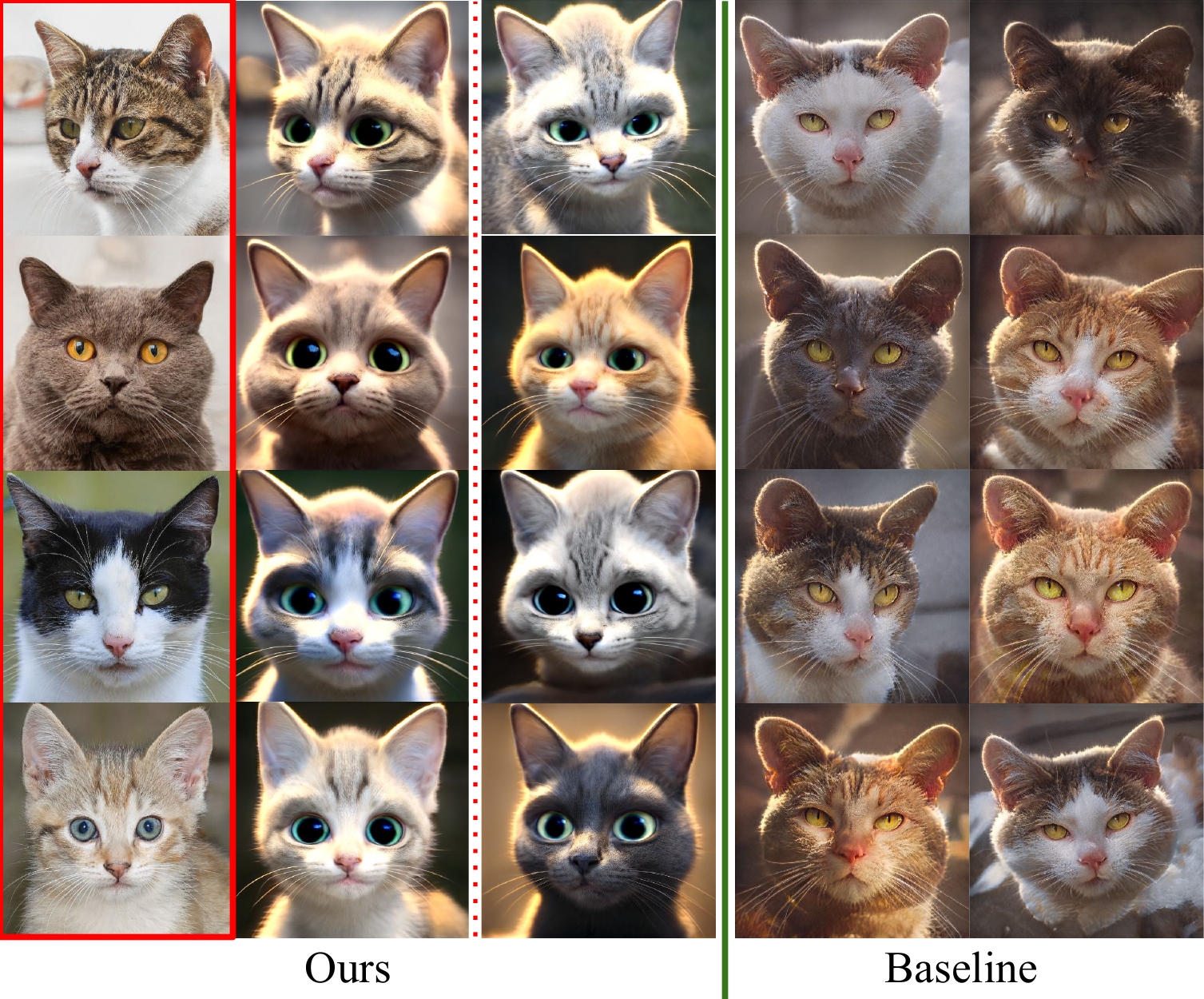}
  \caption{Compare our method with the baseline on "AFHQ-Cat to 3D-stylized cat". The text prompt is long and complicated, describing a complicated appearance, lighting, and rendering style. Our method generated significantly more realistic and natural results than the baseline. }
  \label{fig:nada_cat}
\end{figure}

Similar results are observed on other models and prompts. We experiment with adapting a StyleGAN2-Cat\cite{karras2019analyzing} with a long prompt that contains multiple constraints, including rendering engine, 3D style, texture, and lighting. \cref{fig:nada_cat} shows generated images from our method and the baseline. The baseline model does not properly follow the text description and fails in many aspects. The shading is less realistic than ours, with lots of undesired textures. Our model has more cinematic-like lighting, as described by the prompt. Textures in our results are more smooth than the baseline. Unlike the baseline images that look flat, our results achieve a stronger 3D style with high-quality details. We address these issues from two perspectives: 
\begin{itemize}[noitemsep,nolistsep]
\item{CLIP loss functions is known to have a risk of being minimizd in an adversarial manner and it tends to be trapped in a local minimum \cite{liu2021fusedream}. Prior works observe that optimization overcomes
the CLIP loss by adding pixel-level perturbations to
the image\cite{gal2022stylegan}. Such behavior is problematic when the prompt requests realistic renderings and lighting. Any pixel perturbation could be harmful to a high-quality rendering. Our model, however, uses StableDiffision guidance which happens in the model's latent space. The generator and diffusion guidance are separated by an image Encoder, making the generator more likely to be optimized on a semantically meaningful level rather than adversarially on a pixel level}.
\item{Baseline model uses CLIP text encoders which return one single embedding vector for text prompts. The vector space limits the information capacity and forces compression, reducing the embedding quality when the text prompts are long. Thanks to the fact that StableDiffusion uses a sequence of text embeddings and cross-domain attention, the proposed diffusion guidance is more informative, giving it a higher capability to capture multiple key constraints mentioned in the long text prompts. We further evaluate these results quantitatively in the following section.}
\end{itemize}

\subsubsection{Quantitative Evaluation}

This section quantitatively compares the baseline and our method. We conduct experiments adapting an AFHQ-Cat~\cite{DBLP:journals/corr/abs-1912-01865} generator to generate types of animals indicated by prompts, using our approach and the baseline\cite{gal2022stylegan}. 

\cref{fig:animals} shows 4 uncurated generated samples from our method for each animal type and 1 sample from the baseline. Our method is able to generate images with much higher visual quality and fidelity in these experiments. We provide more details and large-scale image galleries in the appendix.

\cref{tab:FID_all,tab:clip} quantitatively compares the performance. We evaluate the CLIP image-text matching score to measure image-text faithfulness, LPIPS~\cite{zhang2018unreasonable} score to address image diversity, and FID~\cite{heusel2017gans} scores to show image fidelity. CLIP match scores are calculated by extracting CLIP image and text embeddings and computing their dot-product. Although our models are not trained to minimize CLIP loss, our models achieve competitive CLIP and better LPIPS scores than baseline, whereas outperform it in FID scores by a large margin. 
\begin{table}[t]
    \caption{FID scores of Cat/Dog-to-Animals. Ground-truth images are extracted from the AFHQ dataset~\cite{DBLP:journals/corr/abs-1912-01865}. Our models achieve significantly better FIDs than the baseline.}
    \label{tab:FID_all}
    \centering
    \scalebox{1}{
    \begin{tabular}{lcccc}
    \toprule
    \multicolumn{1}{l}{} & \multicolumn{2}{c}{Cat} & \multicolumn{2}{c}{Dog} \\
    \cmidrule(lr){2-3} \cmidrule(lr){4-5}  
    & Ours & NADA & Ours  & NADA \\
    \hline
    Dog/Cat  & \textbf{150.76} & 206.93 & \textbf{124.72}  & 139.35 \\
    Fox  & \textbf{51.51} & 90.40 & \textbf{61.10} & 129.58 \\
    Lion & \textbf{30.34}  & 153.82 & \textbf{52.52}  & 173.81\\
    Tiger & \textbf{19.29}  & 115.46 & \textbf{31.15}  & 223.33\\
    Wolf & \textbf{45.33} & 139.66 & \textbf{71.29}  & 160.00\\
    \bottomrule
    \end{tabular}}
\end{table}

\begin{table}[h!]
    \caption{CLIP and LPIPS score of animal experiments. Our models achieve competitive CLIP scores with better LPIPS scores.}
    \label{tab:clip}
    \centering
    \scalebox{1}{
    \begin{tabular}{lcccc}
    \toprule
    \multicolumn{1}{l}{} & \multicolumn{2}{c}{Ours} & \multicolumn{2}{c}{NADA} \\
    \cmidrule(lr){2-3} \cmidrule(lr){4-5}  
    & CLIP $\uparrow$ & LPIPS $\uparrow$ & CLIP $\uparrow$ & LPIPS $\uparrow$ \\
    \hline
    Dog & {\bf 0.507} & {\bf 0.521} & 0.279 & 0.294 \\
    Hamster & 0.294 & {\bf 0.433} & {\bf 0.319} & 0.421 \\
    Badger & 0.304 & {\bf 0.448} & {\bf 0.324} & 0.417 \\
    Fox & 0.301 & {\bf 0.500} & {\bf 0.305} & 0.490 \\
    Otter & 0.297 & {\bf 0.508} & {\bf 0.335} & 0.349 \\
    Lion & {\bf 0.304} & {\bf 0.458} & 0.302 & 0.412 \\
    Bear & 0.293 & {\bf 0.448} & {\bf 0.305} & 0.393 \\
    Pig & {\bf 0.310} & 0.308 & 0.294 & {\bf 0.532} \\
    \bottomrule
    \end{tabular}}
\end{table}

Such performance gains increase quickly as the text prompts grow longer. To provide quantitative evidence in the long-prompt experiments, we evaluate how well our method and baseline capture the keywords in long long-prompts. Specifically, we calculate the CLIP~\cite{radford2021learning} image-text matching score between the generated image and keywords. We test our method and baseline in adapting the StyleGAN2-Face and Cat generators into generating images with a 3D rendering style indicated by a long text prompt. The prompt contains diverse constraints including eyes, texture, lighting, and style. We run both methods for 2000 iterations after which we sample 2000 images from each model. We split the long text prompt into multiple shorter pieces based on keywords and calculate the average CLIP matching score between generated images and each of them. \cref{tab:clip_key} shows quantitatively that our method works better in capturing key constraints from a long text prompt. 

\begin{table}[h!]
    \caption{CLIP matching score between each keyword within the long prompt and generated images. Our model outperforms the baseline in image-text fidelity in these long-prompts experiments by a large margin.}
    \label{tab:clip_key}
    \centering
    \resizebox{1\linewidth}{!}{
    \begin{tabular}{lcccc}
    \toprule
     & \multicolumn{2}{c}{Face} & \multicolumn{2}{c}{Cat} \\
    \cmidrule(lr){2-3} \cmidrule(lr){4-5}  
    {Prompt Keywords} & Ours & NADA & Ours & NADA\\
    \hline
    3d cute face/cat & {\bf 0.303} & 0.301 & {\bf 0.315} & 0.307\\
    closeup cute and adorable & {\bf 0.247} & 0.222 & {\bf 0.256} & 0.242\\
    cute big circular reflective eyes & {\bf 0.276} & 0.268 & {\bf 0.276} & 0.267\\
    Pixar render & 0.280 & {\bf 0.282} & {\bf 0.251} & 0.247\\
    unreal engine & {\bf 0.273} & 0.271 & {\bf 0.264} & 0.262\\
    cinematic smooth & {\bf 0.227} & 0.208 & {\bf 0.221} & 0.217\\
    intricate detail & {\bf 0.213} & 0.208 & 0.213 & {\bf 0.218}\\
    cinematic lighting & {\bf 0.231} & 0.210 & {\bf 0.230} & 0.229\\
    \bottomrule
    
    \end{tabular}}
\end{table}

\subsection{Timestep Range and Layer Selection}
\label{T and layer exp}

\begin{figure}
\centering
\includegraphics[width=1.0\linewidth]{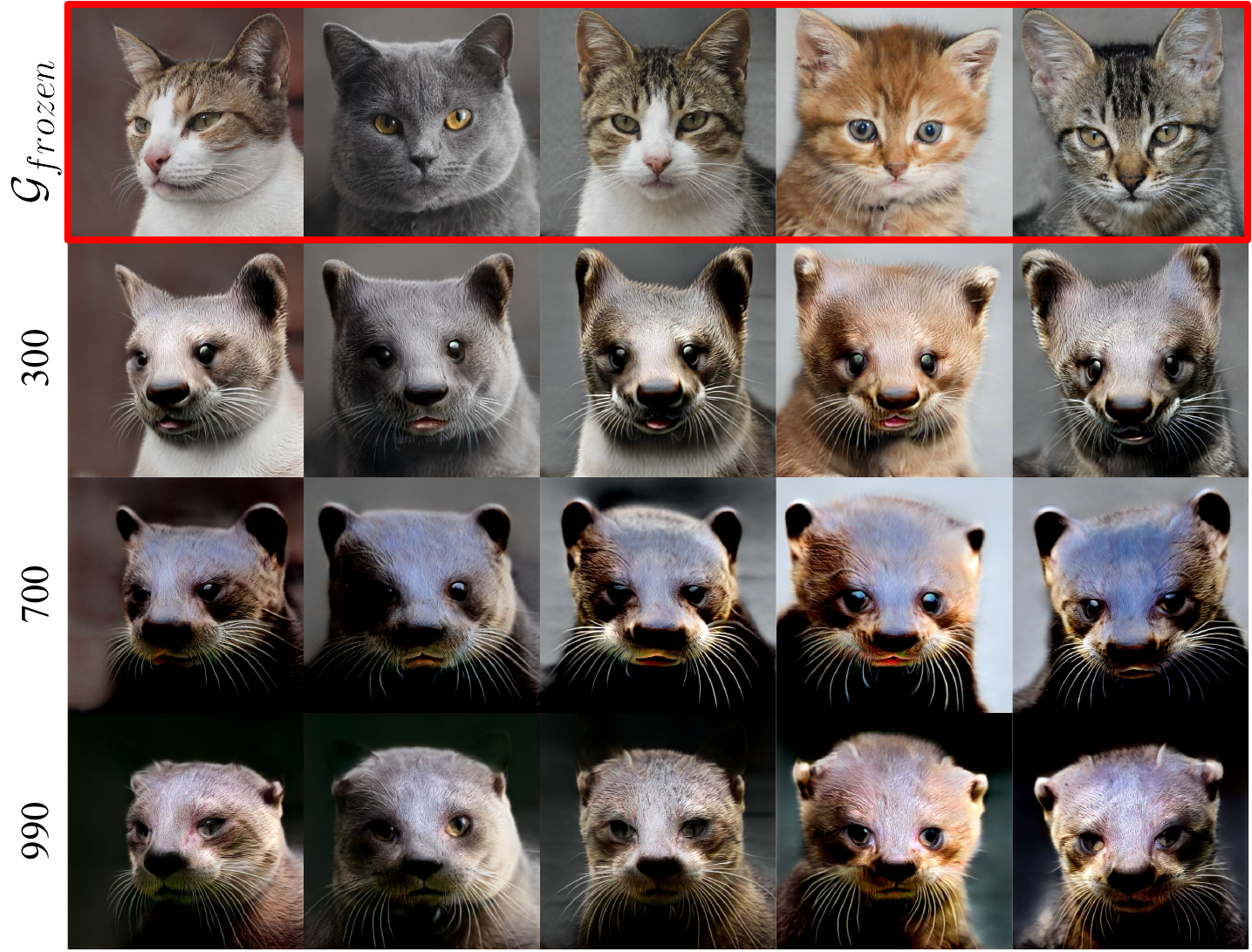}
  \caption{Denoising timestep range ($t_{SDS}$) and model behavior. We fix $T_{min}=0$ and show images from three fine-tuned generators with different $T_{max}$ settings. A larger range enables structure changes and increases image fidelity to the target domain, whereas a smaller range focuses on local changes and prefers faithfulness to the source domain.}  
  \label{fig:T}
\end{figure}

\begin{figure*}
\centering
\includegraphics[width=1\linewidth]{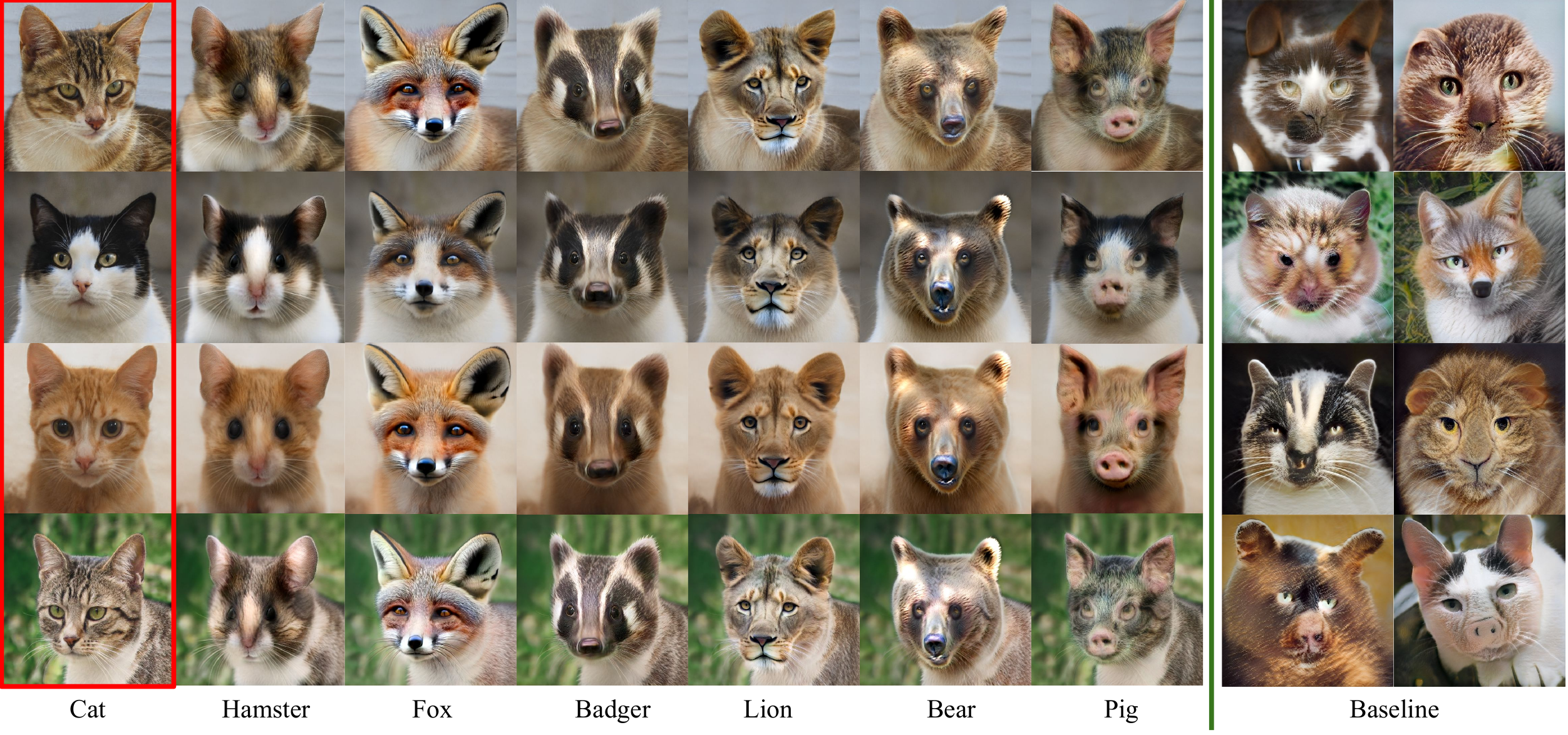}
  \caption{Uncrated samples from our method on Cat-to-8-animals (left). We show results for dog in \cref{fig:results_all} and otter in \cref{fig:T}. For each animal type, we show 1 uncurated sample from baseline (right). Notice the significant visual quality differences. Baseline results are optimized to minimize CLIP metric in an adversarial manner \cite{liu2021fusedream} and have much lower visual quality. Our method generated more visually realistic and natural results, including undistorted facial components, cleaner backgrounds, diverse poses, and higher pose faithfulness. 
  }
  \label{fig:animals}
\end{figure*}

Recall we sample a timestep $t$ from range $T_{SDS}=(T_{min},T_{max})$, in each iteration, based on which we apply the $q$ sample process and update the generator. We experiment with the influence of its configuration on our model behavior. We observe that a large $T_{SDS}$ enables more global structural modifications, while a small $T_{SDS}$ only allows local detail modifications. \cref{fig:T} shows an example of adapting a cat generator to an otter generator. Otter has much smaller ears than cats. Notice the Otter's ears become smaller and more realistic as $T_{SDS}$ increases. This property adds more controllability to the optimization process. A smaller value will be more appropriate when we prefer fidelity to the original domain, and a larger value will be better when the authenticity of the target domain is more important. We include quantitative evaluations in the supplementary material. 

\begin{figure}
\centering
\includegraphics[width=1.0\linewidth]{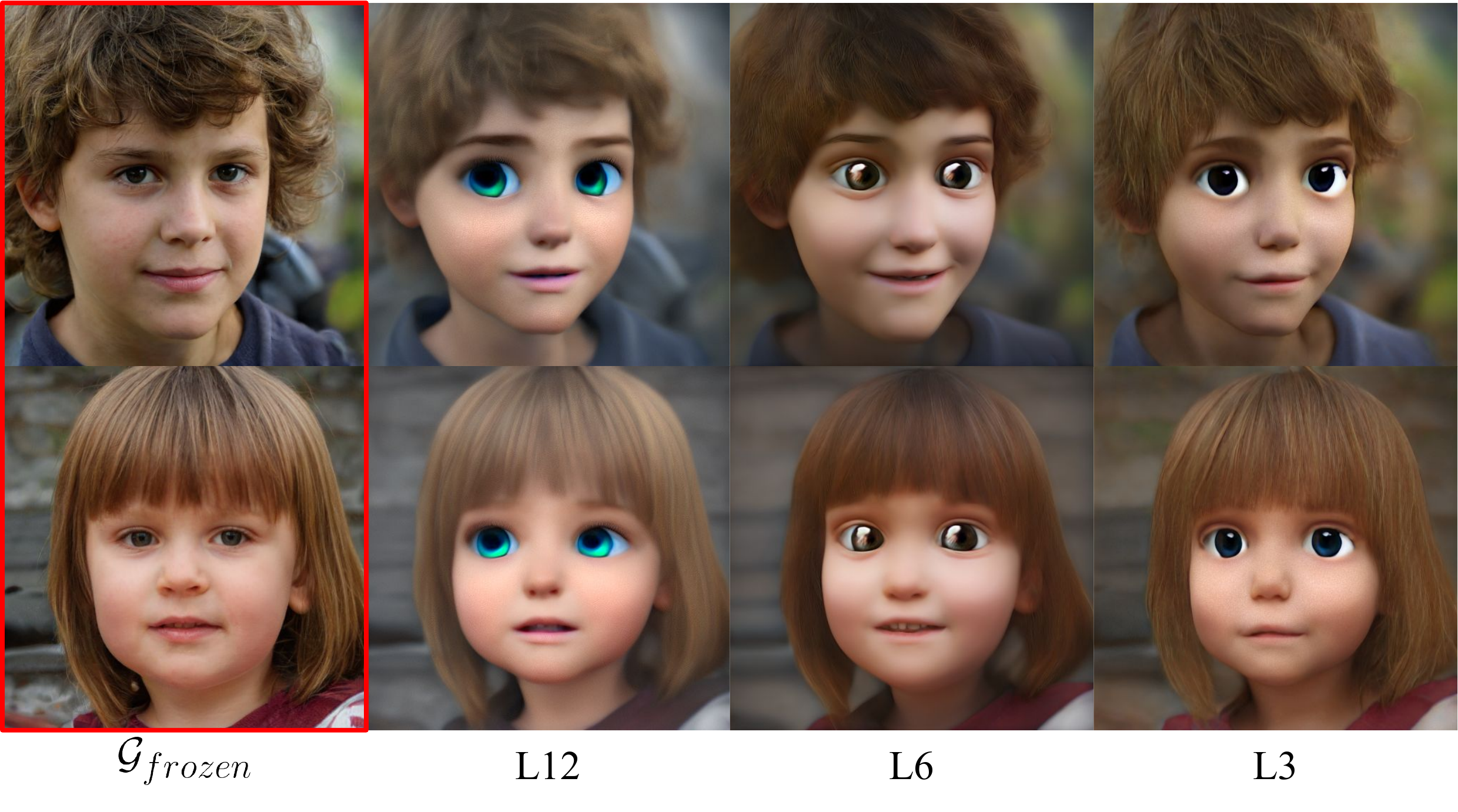}
  \caption{Layer selection reduces blur. We select 12/6/3 layers to optimize for each iteration. Selecting fewer layers requires more iterations of training and we show results with the best visual quality for each layer configuration. With fewer layers selected, the blurry vagueness disappears and hair details are better preserved. 
  }
  \label{fig:layer}
\end{figure}

\subsection{Directional and Reconstruction Regularizer}

This section evaluates the effect of the diffusion-guidance directional and reconstruction regularizer. All experiments in this section use $T_{SDS}=(0,500)$. 
\begin{figure}
\centering
\includegraphics[width=1\linewidth]{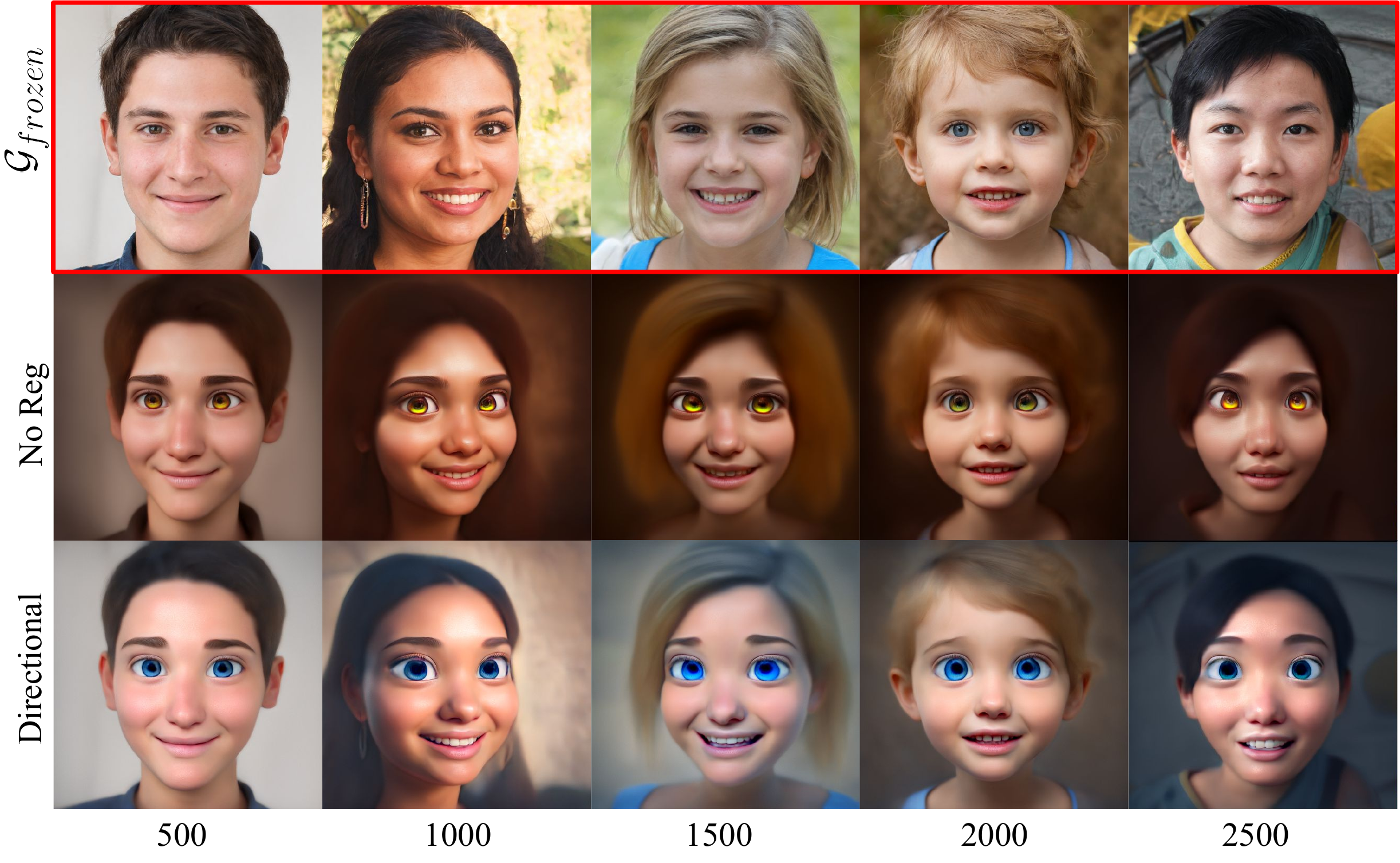}
  \caption{The effect of directional regularizer on FFHQ, as training iteration grows. Notice, $\mathcal{L}_{SDS}^{dir}$ better preserves the details including facial expressions, earrings, and background color, whereas the non-reg approach eventually ignores them.}
  \label{fig:reg_2}
\end{figure}

\cref{fig:reg_2} Compares directional regularizer with a non-regularized baseline. The text prompt is the same as \cref{fig:hero}, requesting a 3D-stylized cute anime face with circular reflective eyes. Notice the earrings and facial expressions are preserved in our approach. Given enough training iterations, the non-reg approach degrades to one single image ignoring the input $\mathbf{z}$, and becomes brownish and dark. Directional regularizer is able to depict the requested style pattern quickly and helps alleviate model collapse. Our generated images have much more faithful facial expressions, color, and diversity.

\begin{figure}
\centering
\includegraphics[width=1\linewidth]{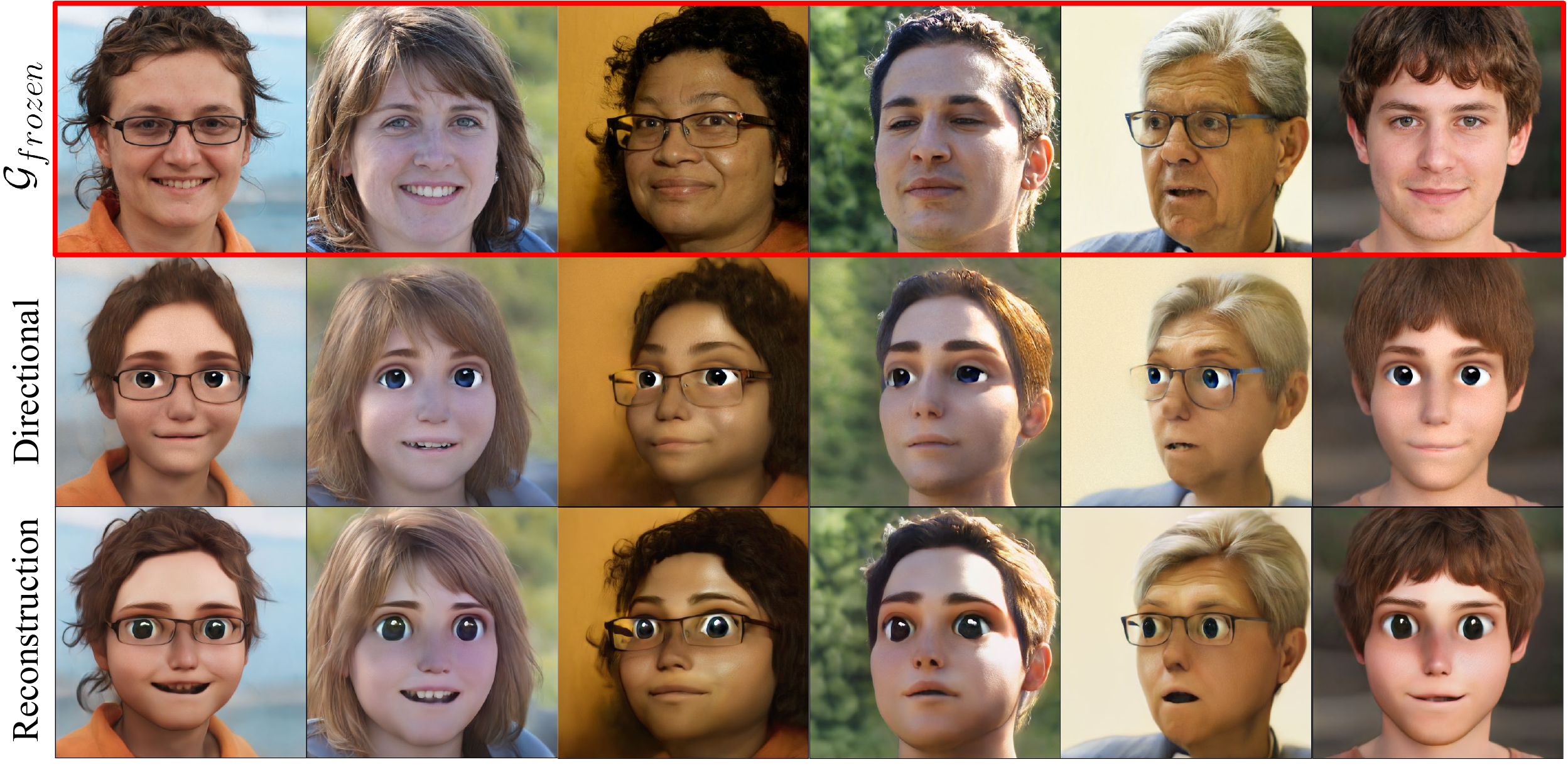}
  \caption{Compare directional and reconstruction regularizers. $\mathcal{L}_{SDS}^{rec}$ is a stronger constraint and preserves better details, whereas $\mathcal{L}_{SDS}^{dir}$ allows new colors to be added such as blue eyes.}
  \label{fig:reconstruction}
\end{figure}

\cref{fig:reconstruction} compares directional and reconstruction regularizer. 
$\mathcal{L}_{SDS}^{rec}$ is a stronger regularizer and keeps most hair details intact. However, it also disables color changes requested by text prompts. $\mathcal{L}_{SDS}^{dir}$ regularizer allows more color modification at a cost of fewer details. 

\subsection{Extension to 3D-Aware Generative Models}
\label{sec:3D_ext}
\begin{figure}[t!]
\centering
\includegraphics[width=1\linewidth]{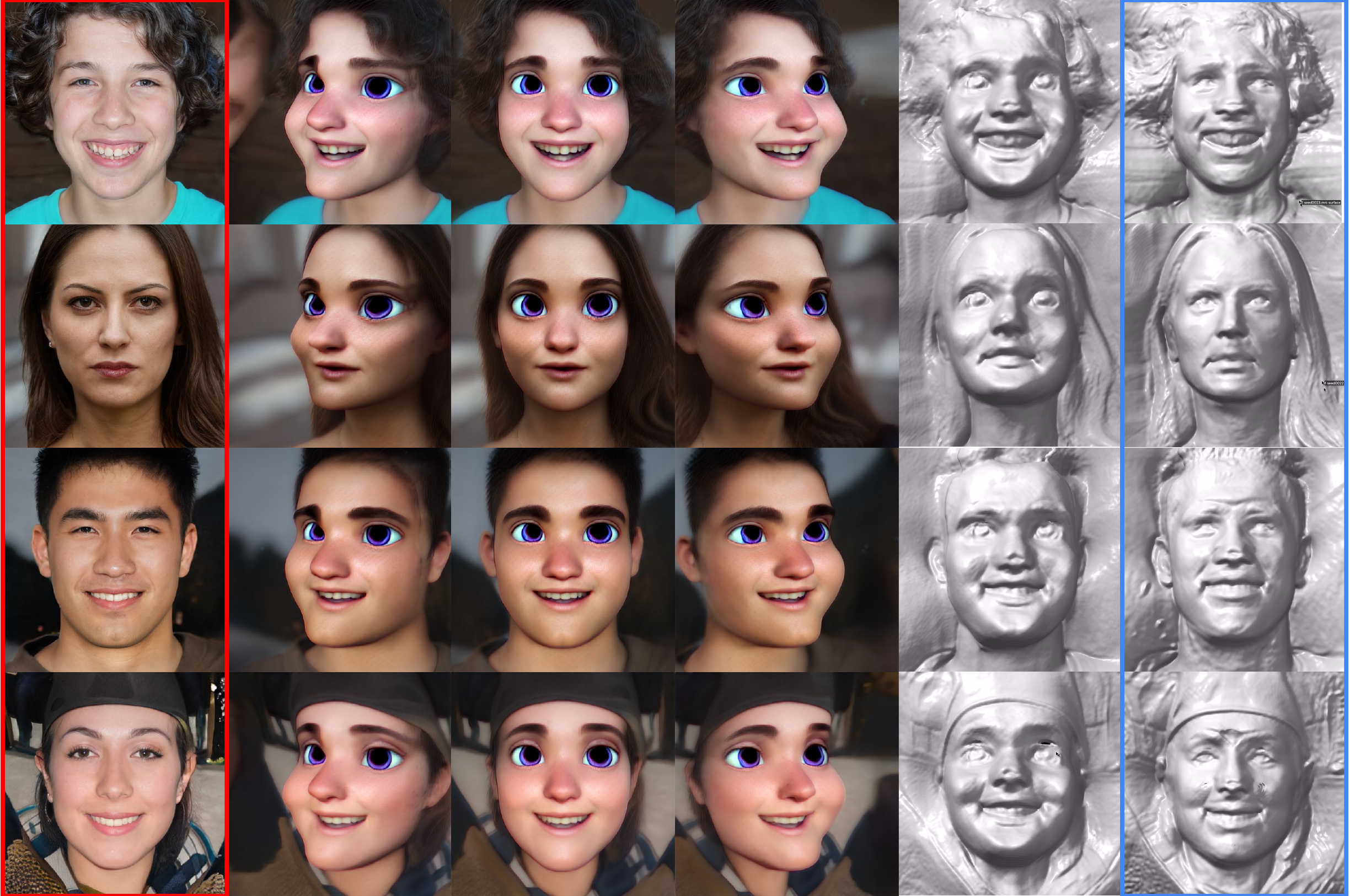}
  \caption{3D domain adaptation on EG3D-Face~\cite{chan2022efficient}. Results from $\mathcal{G}_{train}$ (middle), $\mathcal{G}_{frozen}$ (\textcolor{red}{red} box) and its mesh (\textcolor{blue}{blue} box).}
  \label{fig:3D_face}
\end{figure}

\begin{figure}
\centering
\includegraphics[width=1\linewidth]{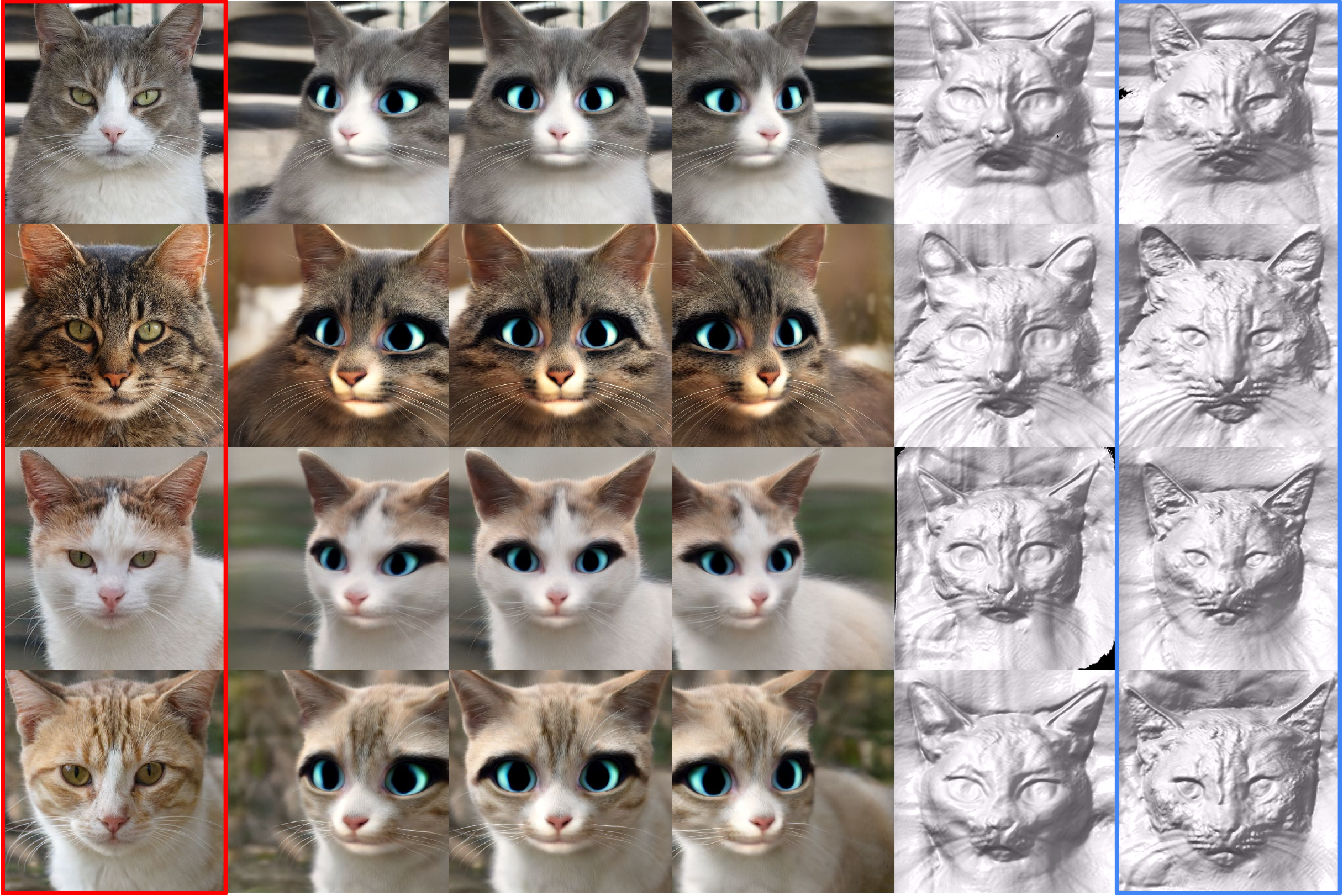}
  \caption{3D domain adaptation on EG3D-Cat~\cite{chan2022efficient}. Results from $\mathcal{G}_{train}$ (middle), $\mathcal{G}_{frozen}$ (\textcolor{red}{red} box) and its mesh (\textcolor{blue}{blue} box).}
  \label{fig:3D_cat}
\end{figure}

We extend our method to 3D Geometry-aware generators from EG3D~\cite{chan2022efficient} on the face and cat models provided by its authors. During optimization, all parameters are frozen except the weights of Conv layers in the tri-plane generator. We add LPIPS ~\cite{zhang2018unreasonable} to the loss function to enforce stronger faithfulness to $\mathcal{G}_{train}$. \cref{fig:3D_face} and \cref{fig:3D_cat} shows our results using the same text prompt in \cref{fig:hero}. The mesh geometries are smoother with larger eye areas as requested by the text prompt. For more results please visit our~\href{https://styleganfusion.github.io/}{\color{urlcolor}{Project Webpage}}.

\subsection{Extension to DreamBooth Guidance}
\label{sec:dreambooth_ext}
\begin{figure}
\centering
\includegraphics[width=1\linewidth]{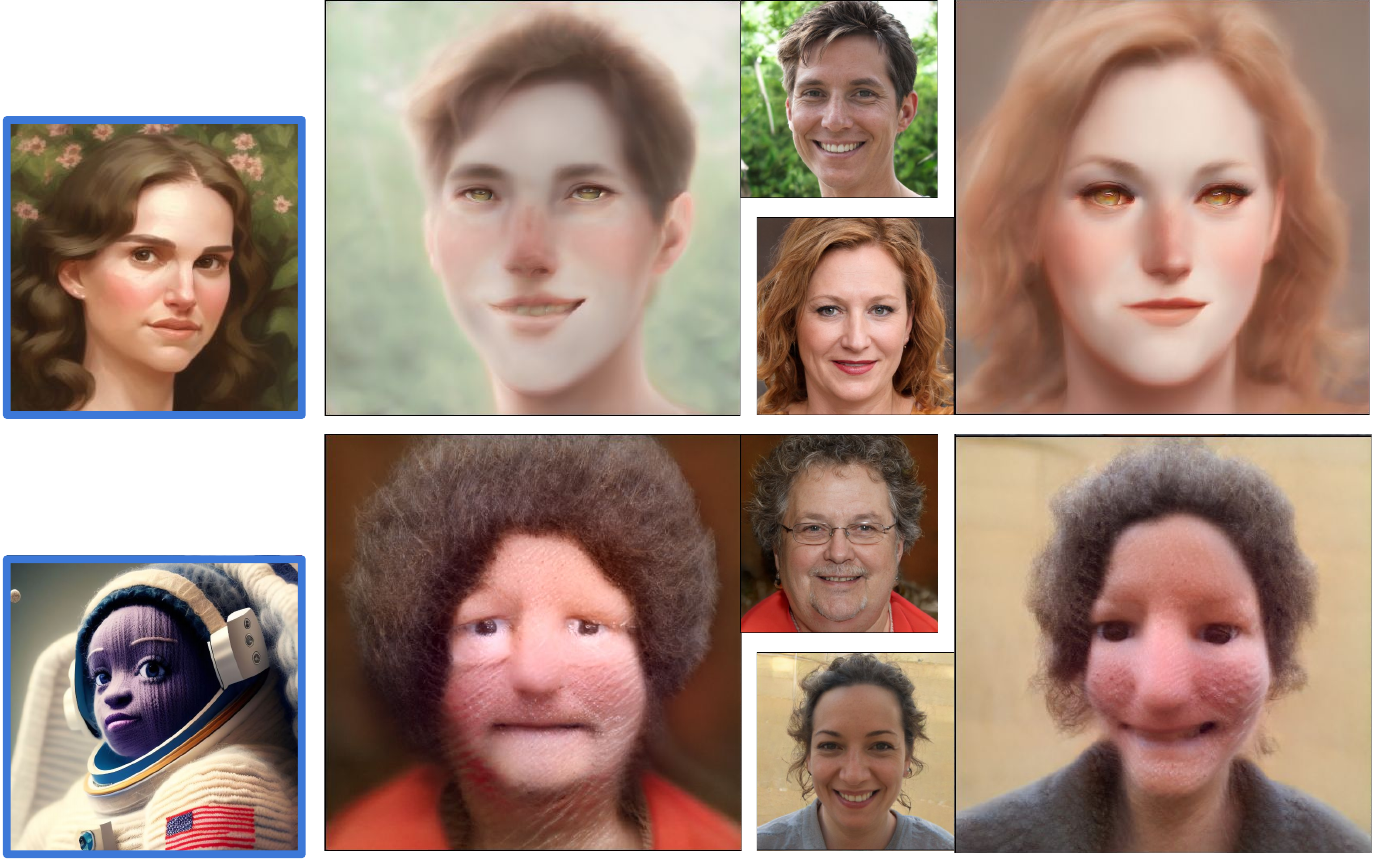}
  \caption{Results from DreamBooth guided models. We show the original DreamBooth StableDiffusion samples in \textcolor{blue}{blue} boxes.}
  \label{fig:dream}
\end{figure}

We additionally extend our method to DreamBooth~\cite{ruiz2022dreambooth} where the StableDiffusion model is finetuned on a few personalized images. We tried public available DreamBooth checkpoints ``Wa-vy'' style~\cite{dreambooth1} and ``Woolitize'' style~\cite{dreambooth2}. \cref{fig:dream} shows our results using the text prompt ``wa-vy style painting close up face'' and ``woolitize close up face''.

\section{Conclusion}
We presented a novel domain adaptation method for image generators that uses StableDiffusion guidance and Score Distillation Sampling. Our method allows flexible control of the magnitude of modifications by selecting the value of $T_{SDS}$. With the introduced diffusion-guidance directional regularizer and layer selection techniques, our model is able to shift the generator to generate new images from a target domain indicated by the text prompt, with improved quality compared to existing methods. We also show that our method can be extended to 3D-aware style-based generators and used with DreamBooth models as guidance.

\clearpage
\newpage
{\small
\bibliographystyle{ieee_fullname}
\bibliography{egbib}
}

\appendix
\section{Full-Text Prompts} 
\label{prompts}
We provide full-text prompts used in the experiments mentioned in our paper:
\begin{enumerate}[noitemsep,nolistsep]
    \item ``3d human face, closeup cute and adorable, cute big circular reflective eyes, Pixar render, unreal engine cinematic smooth, intricate detail, cinematic''
    \item ``3d cat, closeup cute and adorable, cute big circular reflective eyes, Pixar render, unreal engine cinematic smooth, intricate detail, cinematic''
    \item ``Joker''
    \item ``High quality 3 d render of very fluffy cat, highly detailed, unreal engine cinematic smooth, in the style of detective Pikachu blade runner, hannah yata charlie immer, neon light, low angle, uhd 8 k, sharp focus''
    \item ``An epic fantasy comic book style portrait painting of dog, very expressive, light blue piercing eyes, round face, character design by mark ryden and pixar and hayao miyazaki, unreal 5, daz, hyperrealistic, octane render, cosplay, rpg portrait, dynamic lighting, intricate detail, summer vibrancy, cinematic''
    \item ``Werewolf''
    \item ``Photo of a dog/hamster/badger/fox/otter/lion/bear/pig''
    \item ``Cinematic portrait of brutal epic dark dog, concept art, artstation, glowing lights, highly detailed''
    \item ``Photo of a car, TRON wheel''
    \item ``Render of a car mod for GTA San Andreas, game screenshot''
    \item ``Sketch of a car, pen and ink sketch''
    \item ``A masterpiece ultrarealistic ultradetailed portrait of a incredibly beautiful human face, baroque renaissance, in the night forest. medium shot, intricate, elegant, highly detailed. trending on artstation, digital art, by stanley artgerm lau, wlop, rossdraws, james jean, andrei riabovitchev, marc simonetti, yoshitaka amano. background by james jean and gustav klimt, light by julie bell, 4 k, porcelain skin. 
    \item ``Very beautiful portrait of an extremely cute and adorable face, smooth, perfect face, fantasy, character design by mark ryden and pixar and hayao miyazaki, sharp focus, concept art, harvest fall vibrancy, intricate detail, cinematic lighting, hyperrealistic, 3 5 mm, diorama macro photography, 8 k, 4 k''
    \item ``Charcoal pencil sketch of human face, lower third, high contrast, black and white''
    \item ``A very beautiful anime girl, full body, long braided curly silver hair, sky blue eyes, full round face, short smile, casual clothes, ice snowy lake setting, cinematic lightning, medium shot, mid-shot, highly detailed, trending on Artstation, Unreal Engine 4k, cinematic wallpaper by Stanley Artgerm Lau, WLOP, Rossdraws, James Jean, Andrei Riabovitchev, Marc Simonetti, and Sakimichan''
\end{enumerate}

The prompt IDs used in experiments in main text are:
\begin{itemize}[noitemsep,nolistsep]
    \item \cref{fig:hero}: Face -- prompt 1, Cat -- prompt 2.
    \item \cref{fig:results_all}: (Face) $\rightarrow$ Joker -- prompt 3, (Cat) $\rightarrow$ Pikachu Cat -- prompt 4, (Dog) $\rightarrow$ Comic Dog -- prompt 5, (Face) $\rightarrow$ Werewolf -- prompt 6, (Cat) $\rightarrow$ Dog -- prompt 7, (Dog) $\rightarrow$ Epic Dark Dog -- prompt 8.
    \item \cref{fig:GTA}: (Car) $\rightarrow$ TRON wheel -- prompt 9, (Car) $\rightarrow$ GTA car -- prompt 10, (Car) $\rightarrow$ Sketch -- prompt 11.
    \item \cref{fig:nada_2}, \cref{fig:layer}, \cref{fig:reg_2}, \cref{fig:reconstruction} and \cref{fig:3D_face}: prompt 1.
    \item \cref{fig:nada_cat} and \cref{fig:3D_cat}: prompt 2.
    \item \cref{fig:T} and \cref{fig:animals}: prompt 7.
    \item \cref{tab:FID_all} and \cref{tab:clip}: prompt 7, each run takes one animal type.
    \item \cref{tab:clip_key}: Face -- prompt 1, Cat -- prompt 2.
\end{itemize}

\section{Implementation Details}
\noindent \textbf{Architecture.} We use the StyleGAN2 PyTorch \cite{NEURIPS2019_9015} config-f implementation by \cite{stylegan_ros}. Checkpoint resolutions are: FFHQ $1024\times1024$, AFHQ-Cat/Dog $512\times512$. We adapt the StableDiffusion v1.4\cite{rombach2021highresolution} model from Diffusers \cite{diffusers}. Training code are built upon 
Stable-DreamFusion \cite{Stable-DreamFusion}.

\noindent \textbf{Training.} The latent mapping MLP layer, all ToRGB layers, and bias are frozen and we only update weights in Conv layers. We used Adam optimizer with default parameters and a learning rate of $5\times 10^{-4}$. All models were trained for 2000 training steps with batch size of 1. Training takes about 20 minutes on an A100 GPU with a memory cost of 14.7G for 1024 resolution and 12.5G for 512 resolution. 
We set $w_t=1-\bar{\alpha}_t$ following the implementation in \cite{Stable-DreamFusion}. The training time for each experiment varies from 20 minutes to 2 hours on an NVIDIA V100, with a memory cost of 25G when batch size equals one. In some experiments, we found it helpful to add a small LPIPs loss to increase stability.

\section{More Quantitative Evaluations}
\noindent \textbf{Fréchet Inception Distance.} To further quantitatively measure the quality of generated images, we calculate and compare the clean Fréchet Inception Distance score \cite{DBLP:journals/corr/abs-2104-11222} for the animal experiments. Ground truth images are needed when calculating FID. We use the AFHQ dataset \cite{DBLP:journals/corr/abs-1912-01865} and manually extract ground truth images for Fox/Lion/Tiger/Wolf from its ``wild'' subclass. We use the default batch size of 256 for all FID calculations.

\cref{tab:FID_1} shows the FID scores for Cat-to-Animals. And \cref{tab:FID_2} for Dog-to-Animals. Notice, none of these models are exposed to a single ground truth image and the scores are attained in a \emph{zero-shot} manner. 

Our method achieved \emph{significantly better FID scores} than the baseline: StyleGAN-NADA\cite{gal2022stylegan} in all cat/dog-to-animals experiments. 

\begin{table}[t]
    \caption{FID scores of Cat $\rightarrow$ Animals}
    \label{tab:FID_1}
    \centering
    \scalebox{1}{
    \begin{tabular}{lcccc}
    \toprule
    \multicolumn{1}{l}{} & \multicolumn{3}{c}{\textbf{Ours}} & \multicolumn{1}{c}{\textbf{NADA}} \\
    \cmidrule(lr){2-4}  
    & T750 & T500 & T300  &  \\
    \hline
    Dog  & 165.0258 & 155.1359 & \textbf{150.7622} & {206.9277} \\
    Fox  & 55.1440 & 54.1920 & \textbf{51.5124} & {90.4007} \\
    Lion & 59.4057  & 35.1491 & \textbf{30.3362} & {153.8199}\\
    Tiger & 19.9225  & \textbf{17.0460} & 19.2870 & {115.4611}\\
    Wolf & 66.3091 & \textbf{42.5760} & 45.3286 & {139.6573}\\
    \bottomrule
    \end{tabular}}
\end{table}

\begin{table}[t]
    \caption{FID scores of Dog $\rightarrow$ Animals}
    \label{tab:FID_2}
    \centering
    \scalebox{1}{
    \begin{tabular}{lcccc}
    \toprule
    \multicolumn{1}{l}{} & \multicolumn{3}{c}{\textbf{Ours}} & \multicolumn{1}{c}{\textbf{NADA}} \\
    \cmidrule(lr){2-4} 
    & T750 & T500 & T300  &  \\
    \hline
    Cat  & \textbf{115.1011} & 130.2553 & 124.7166 & {139.3474} \\
    Fox  & 65.3474& 67.3951 & \textbf{61.0955} & {129.5795} \\
    Lion & 67.3678  & 55.3808 & \textbf{52.5201} & {173.8075}\\
    Tiger & \textbf{26.9477}  & 28.1698  & 31.1502 & {223.3331}\\
    Wolf & 133.7088 & 81.0359  & \textbf{71.2905} & {159.9959}\\
    \bottomrule
    \end{tabular}}
\end{table}

\section{$T_{SDS}$ Configurations}
\begin{figure}[ht]
\centering
\includegraphics[width=1\linewidth]{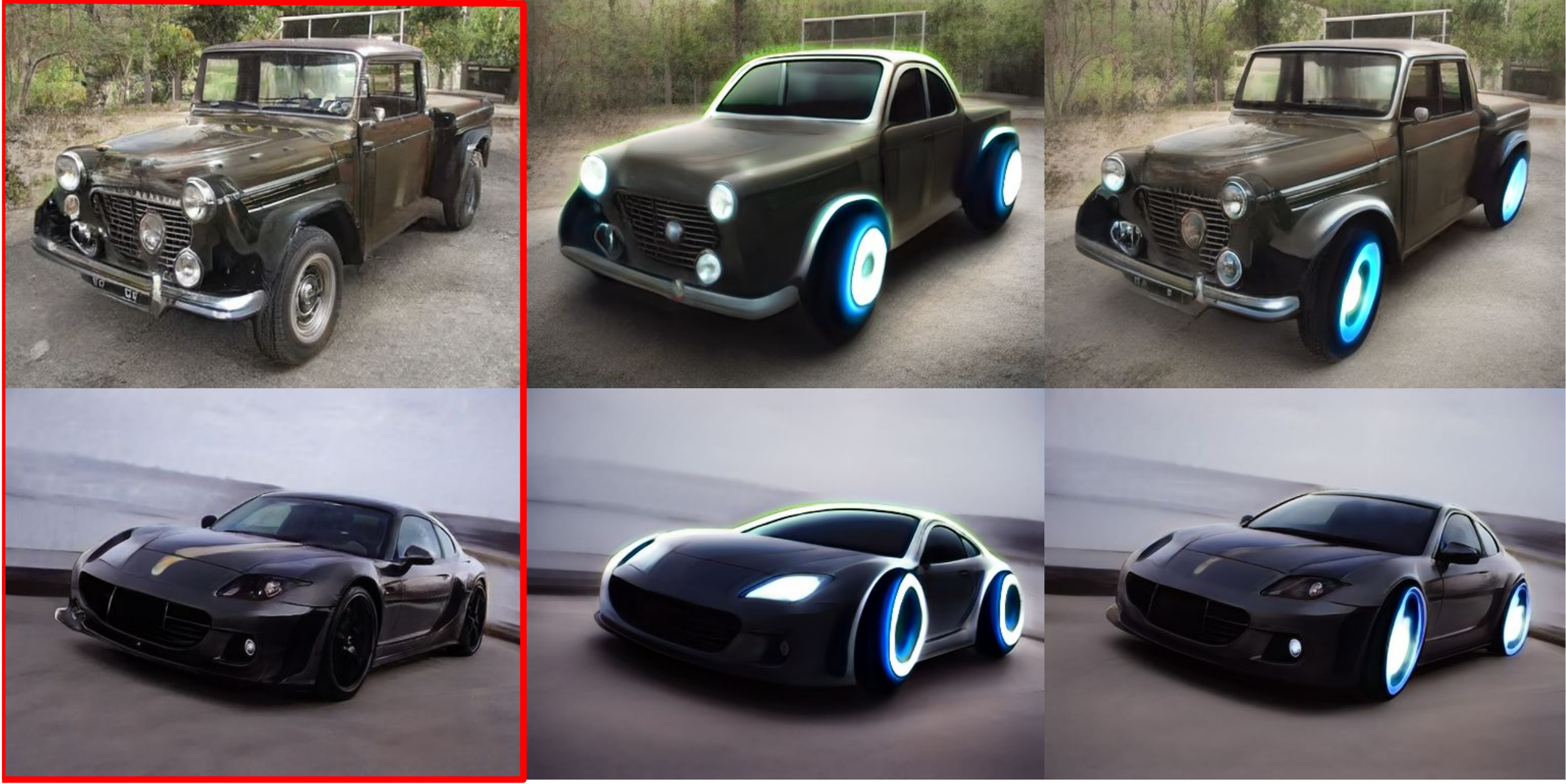}
  \caption{$T_{SDS}$ settings on Car to TRON wheel. 2nd column uses $T_{max}=600$, 3rd column use $T_{max}=300$}
  \label{fig:appendix_9}
\end{figure}
We provide more experiments on the $T_{SDS}$ settings in our model. In \cref{tab:FID_1} and \cref{tab:FID_2}, we compared FID metric for three $T_{SDS}$ configurations. In most cases, $T_{max} = 300$ has a better FID score. \cref{fig:appendix_9} visually shows the its influence. A higher value allows more compositional modifications and a lower value only allows local changes.

\section{Large-Scale Image Galleries}
\begin{figure*}
\centering
\includegraphics[width=0.75\linewidth]{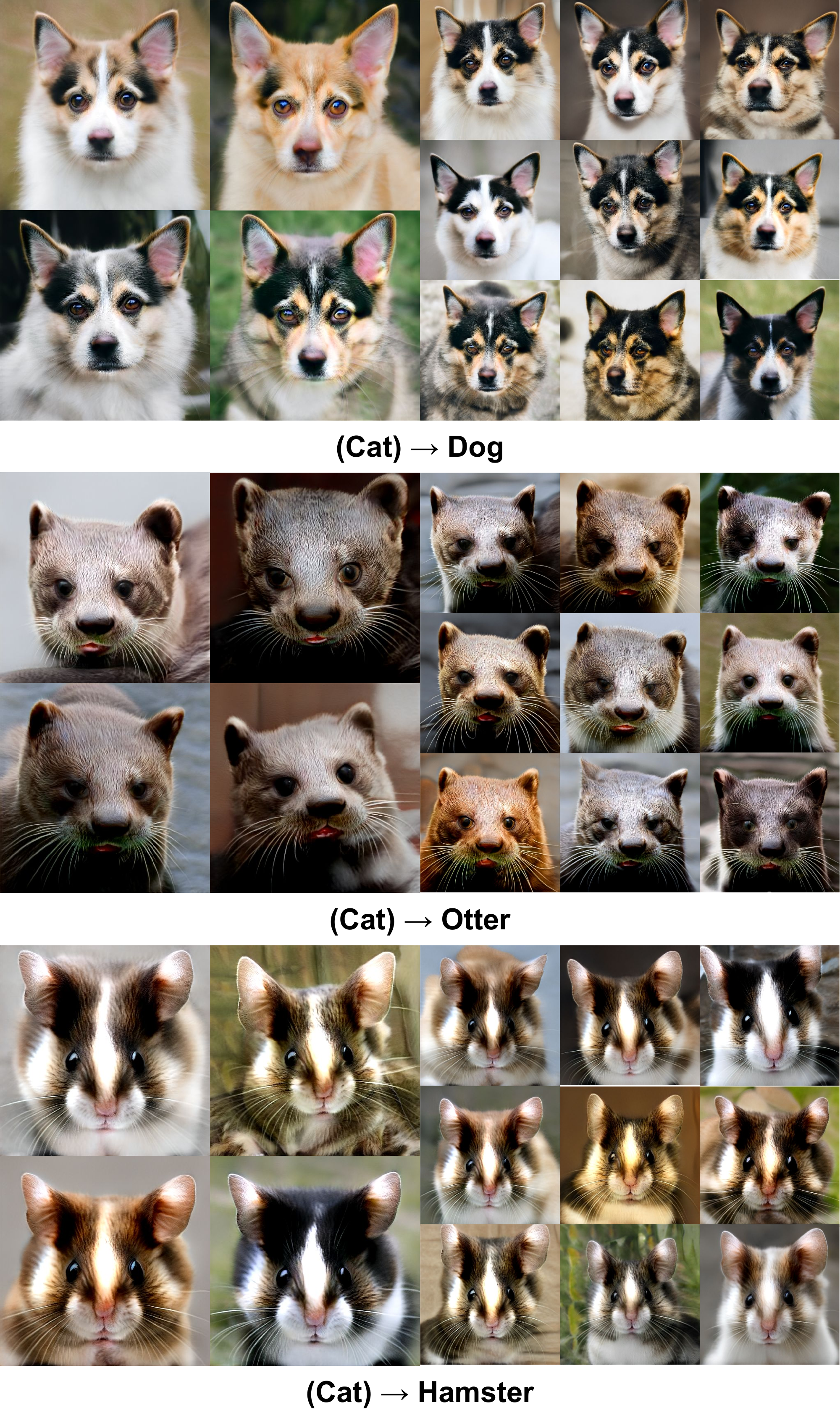}
  \caption{Our results for AFHQ-cat to other animals. }
  \label{fig:appendix_1}
\end{figure*}

\begin{figure*}
\centering
\includegraphics[width=0.75\linewidth]{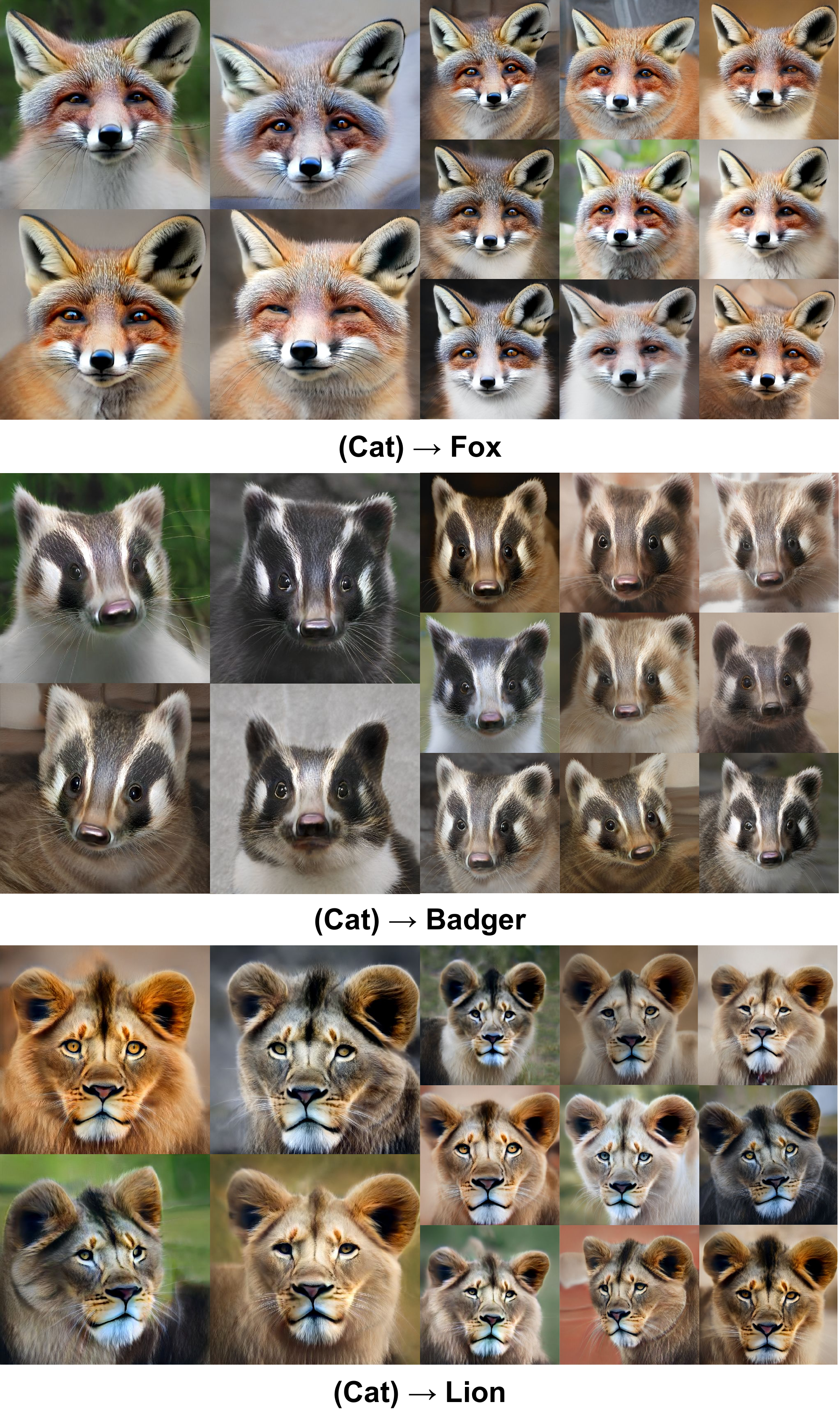}
  \caption{Our results for AFHQ-cat to other animals.}
  \label{fig:appendix_2}
\end{figure*}

\begin{figure*}
\centering
\includegraphics[width=0.75\linewidth]{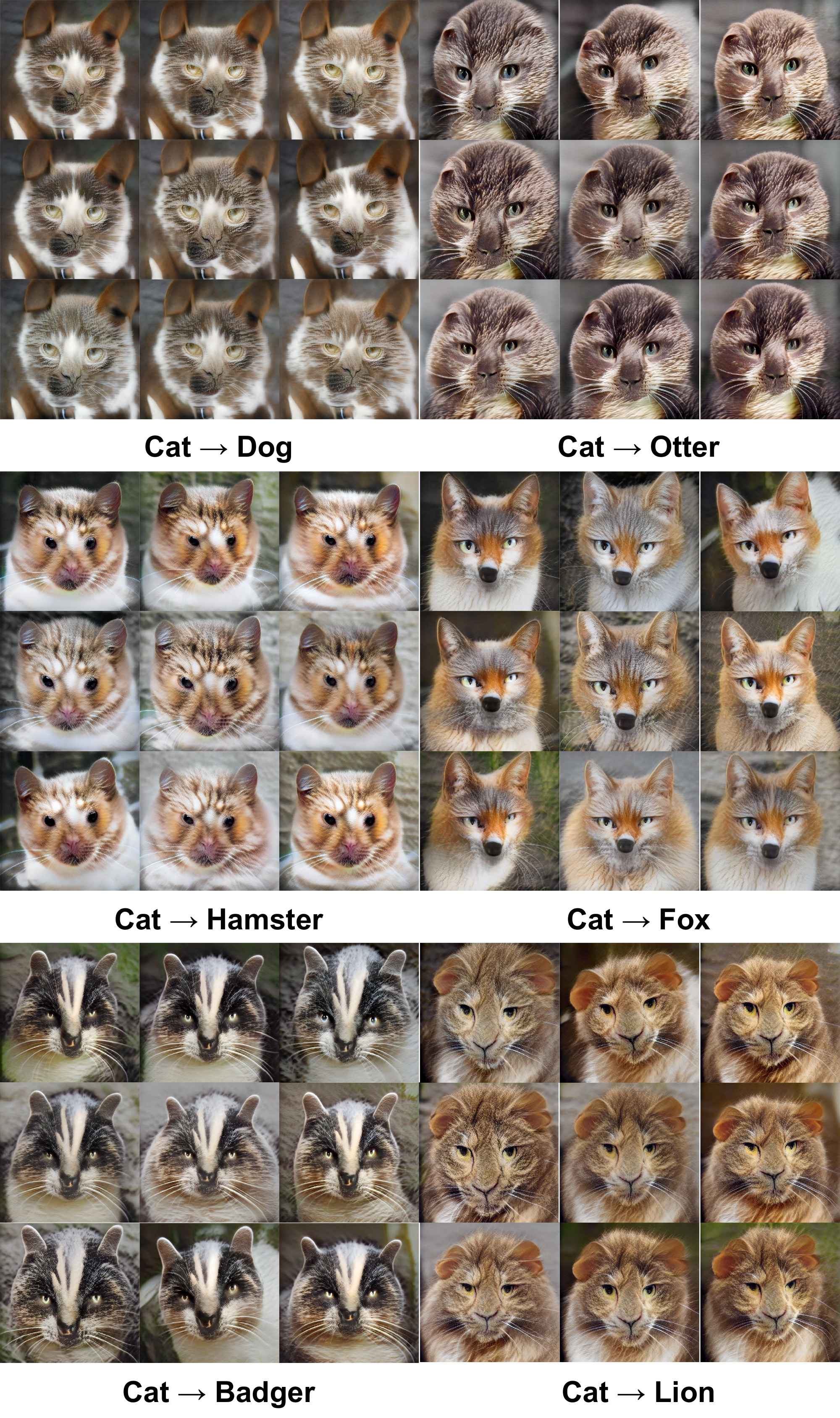}
  \caption{Results from baseline StyleGAN-NADA for AFHQ-Cat to other animals. Baseline has distorted facial components, unrealistic texture, and lower diversity than ours. }
  \label{fig:appendix_3}
\end{figure*}

\begin{figure*}
\centering
\includegraphics[width=0.87\linewidth]{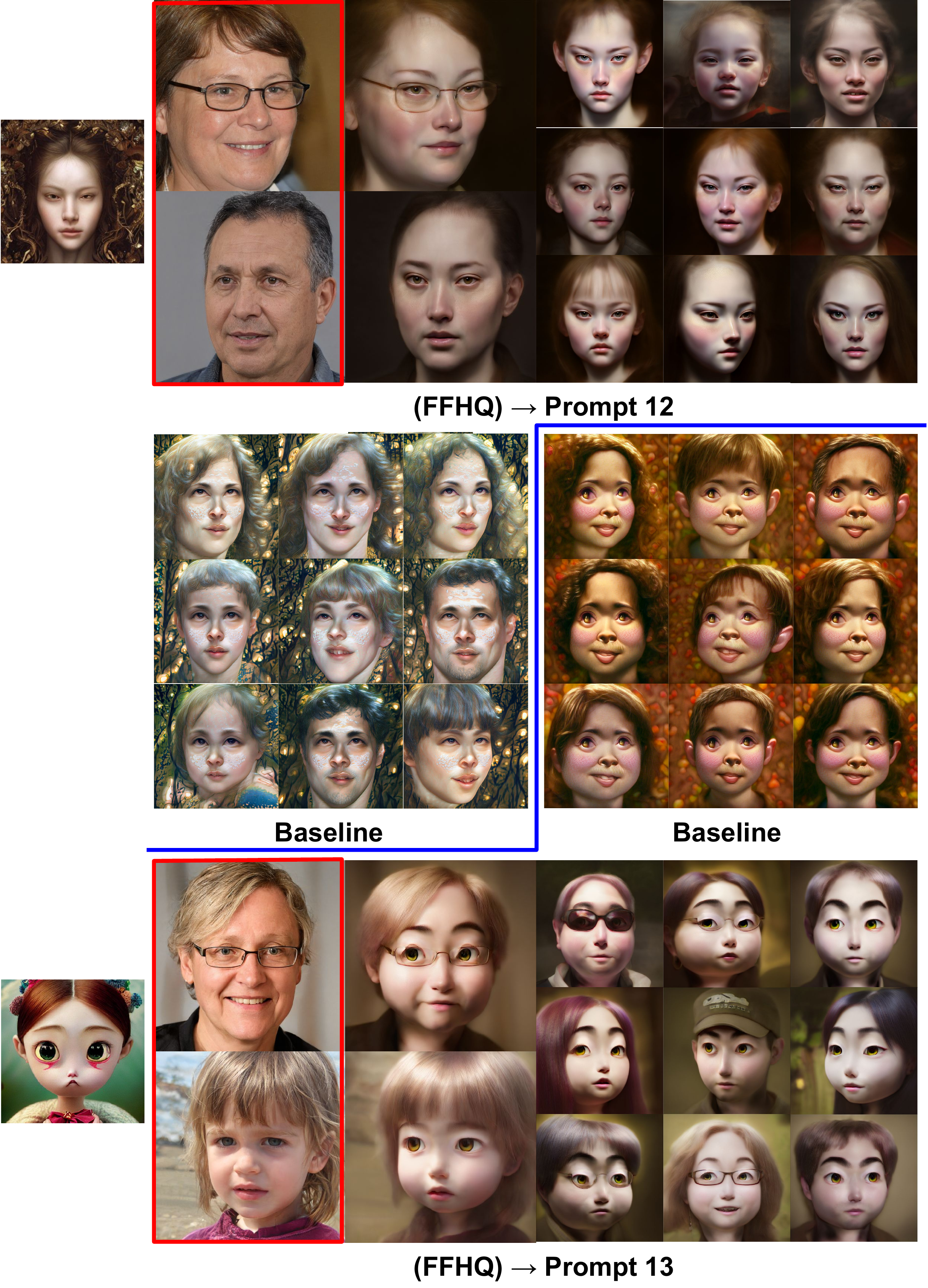}
  \caption{Addtional results from our model and baseline on FFHQ face experiments. Full-text prompts are available in \cref{prompts}. On the left side, we show one sample from StableDiffusion for each prompt. Samples of $\mathcal{G}_{frozen}$ are marked in red boxes.}
  \label{fig:appendix_4}
\end{figure*}

\begin{figure*}
\centering
\includegraphics[width=0.87\linewidth]{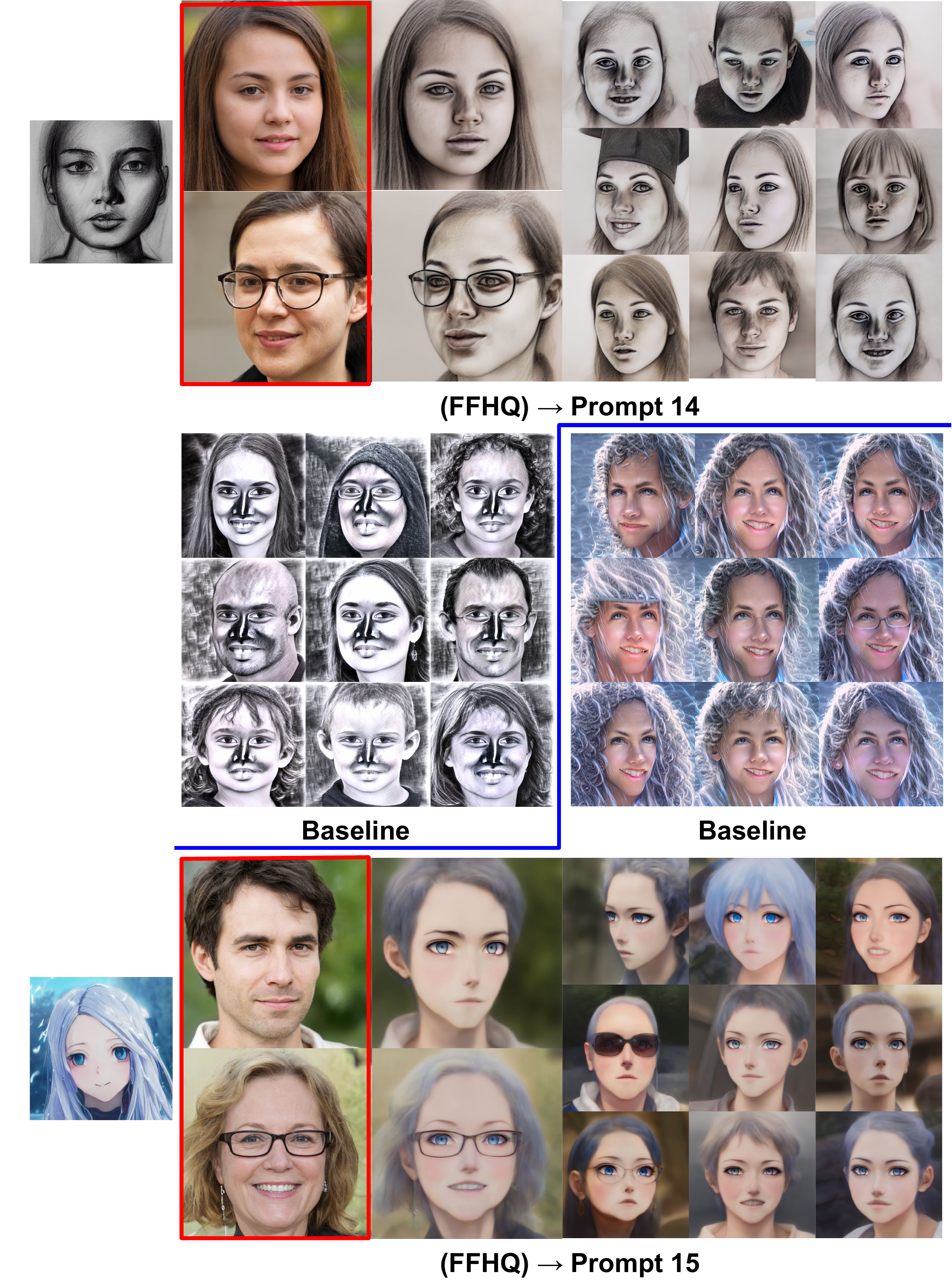}
  \caption{Addtional results from our model and baseline on FFHQ face experiments. Full-text prompts are available in \cref{prompts}. On the left side, we show one sample from StableDiffusion for each prompt. Samples of $\mathcal{G}_{frozen}$ are marked in red boxes.}
  \label{fig:appendix_5}
\end{figure*}

\begin{figure*}
\centering
\includegraphics[width=1\linewidth]{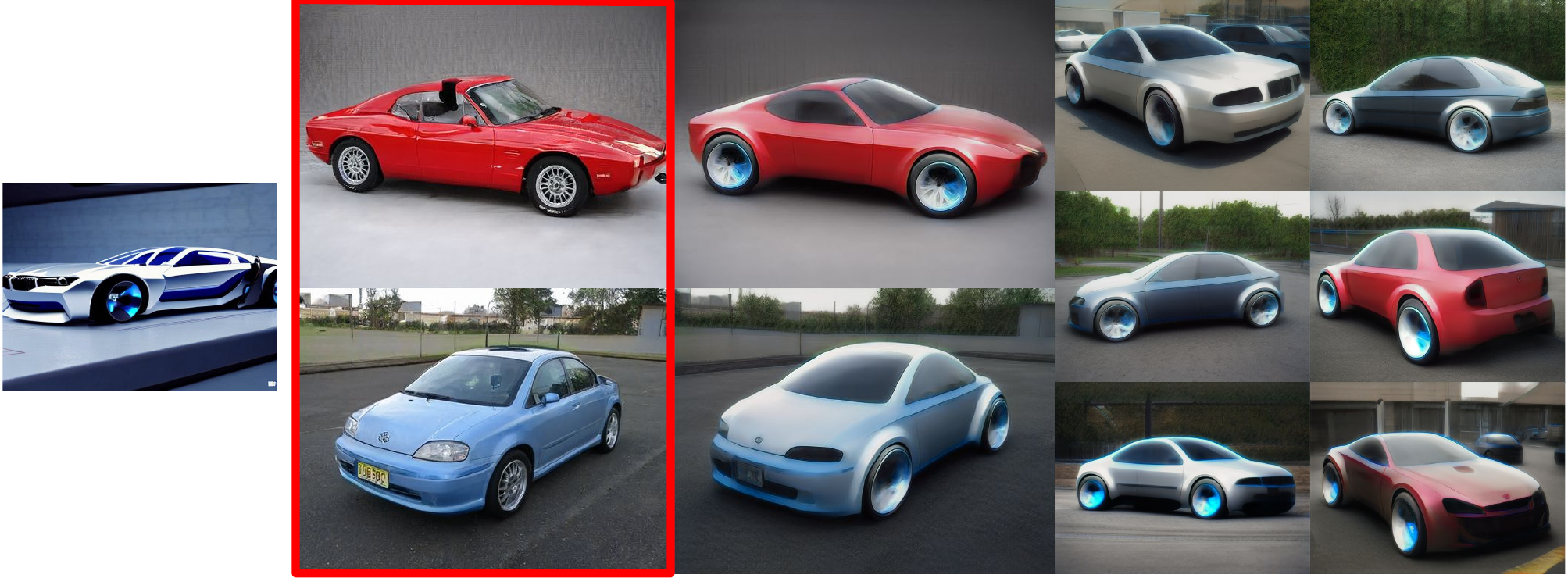}
  \caption{Addtional results from our model on Car to concept car experiments. Full-text prompt: Cyberpunk BMW concept-inspired sports car on the road, futuristic look, highly detailed body, very expensive, photorealistic camera shot, bright studio setting, light reflections, unreal engine 5 quality render. On the left side, we show one sample from StableDiffusion for each prompt. Samples from $\mathcal{G}_{frozen}$ are marked in the red box.}
  \label{fig:appendix_7}
\end{figure*}

\begin{figure*}
\centering
\includegraphics[width=1\linewidth]{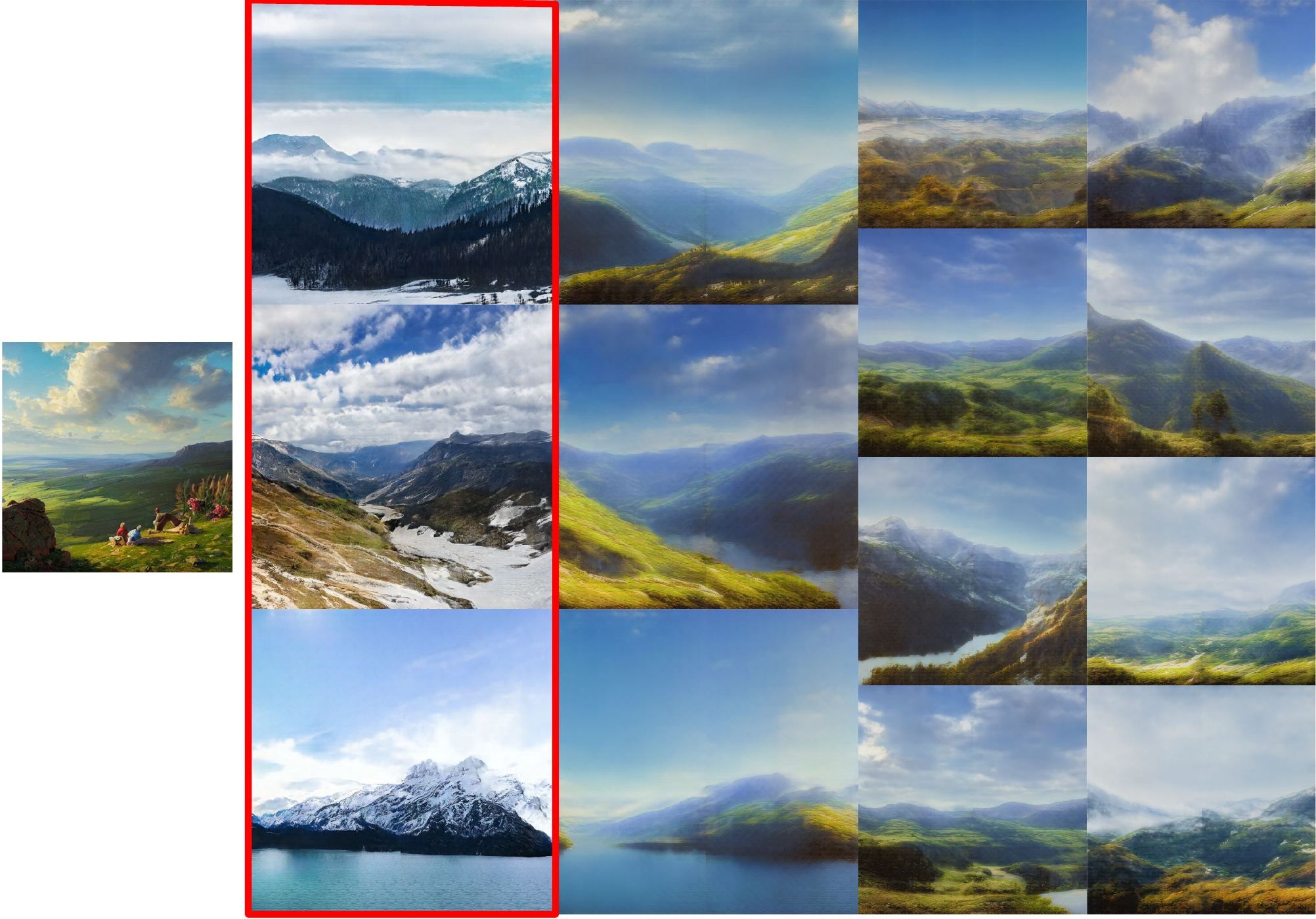}
  \caption{Addtional results from our model on Landscape experiment. Full-text prompt: vast view of landscape by Vladimir Volegov and Alexander Averin and Peder Mørk Mønsted and Adrian Smith and Raphael Lacoste. On the left side, we show one sample from StableDiffusion for each prompt. Samples from $\mathcal{G}_{frozen}$ are marked in the red box.}
  \label{fig:appendix_8}
\end{figure*}

We provide additional visual results for multiple models and prompts. \cref{fig:appendix_1} and  \cref{fig:appendix_2} are generated images from our model for the cat-to-animals experiments. \cref{fig:appendix_3} shows the results from baseline: StyleGAN-NADA. \cref{fig:appendix_4} and \cref{fig:appendix_5} are our results from FFHQ face model. Additional results for car and landscape are shown in \cref{fig:appendix_7} and \cref{fig:appendix_8}.

\end{document}